\documentclass{article}

\usepackage{svg}
\usepackage{microtype}
\usepackage{graphicx}
\usepackage{subfig}
\usepackage{booktabs} 
\usepackage{url}
\usepackage{duckuments} 
\usepackage{multirow}
\usepackage{amsmath}
\usepackage{amsfonts}
\usepackage{amsthm}
\usepackage{pifont}
\usepackage{amssymb}
\newcommand{\xmark}{\ding{55}}%
\usepackage[normalem]{ulem}

\RequirePackage{color}

\newcommand{\y}{{\bf y}}
\newcommand{\x}{{\bf x}}
\newcommand{\z}{{\bf z}}
\newcommand{\uu}{{\bf u}}
\newcommand{\s}{{\bf s}}
\newcommand{\vv}{{\bf v}}

\usepackage{xspace}

\newcommand{\alg}{\texttt{MYOW}\xspace}
\newcommand{\wmse}{\texttt{W-MSE}\xspace}

\newcommand{\byol}{\texttt{BYOL}\xspace}
\newcommand{\barlow}{\texttt{BarlowTwins}\xspace}
\newcommand{\simclr}{\texttt{SimCLR}\xspace}

\definecolor{keithpurple}{rgb}{0.41, 0.05, 0.675}

\newcommand{\deltacc}{$\delta$-Acc}

\definecolor{misocolor}{rgb}{0.01, 0.75, 0.24}

\definecolor{rangreen}{rgb}{0,0.39,0}

\usepackage{hyperref}

\makeatletter
\newtheorem*{rep@theorem}{\rep@title}
\newcommand{\newreptheorem}[2]{%
\newenvironment{rep#1}[1]{%
 \def\rep@title{#2 \ref{##1}}%
 \begin{rep@theorem}}%
 {\end{rep@theorem}}}
\makeatother

\newreptheorem{theorem}{Theorem}
\newreptheorem{lemma}{Lemma}
\newreptheorem{proposition}{Proposition}

\usepackage{xcolor,amsmath}
\usepackage[ruled]{algorithm2e}
\DontPrintSemicolon

\SetKwComment{Comment}{\color{green!50!black}// }{}

\newcommand{\FuncCall}[2]{\texttt{\bfseries #1(#2)}}
\SetKwProg{Function}{function}{}{}

\SetKwInOut{Input}{input}
\SetKwProg{Init}{init}{}{}

\DeclareUnicodeCharacter{2212}{-}

\newcommand{\algorithmicrequire}{\textbf{Input:}}
\newcommand{\algorithmicoutput}{\textbf{Output:}}
\newcommand{\algorithmicensure}{\textbf{Initialize:}}

\usepackage[preprint, nonatbib]{neurips_2021}

\usepackage[utf8]{inputenc} 
\usepackage[T1]{fontenc}    
\usepackage{hyperref}       
\usepackage{url}            
\usepackage{booktabs}       
\usepackage{amsfonts}       
\usepackage{nicefrac}       
\usepackage{microtype}      
\usepackage{xcolor}         

\title{Mine Your Own vieW: Self-Supervised Learning Through Across-Sample Prediction}

\author{Mehdi Azabou\\ Georgia Tech \And
Mohammad Gheshlaghi Azar\\ DeepMind \And
Ran Liu\\ Georgia Tech \And
Chi-Heng Lin\\ Georgia Tech \And
Erik C. Johnson\\ JHU-APL \And
Kiran Bhaskaran-Nair\\ WashU-St Louis \And
Max Dabagia\\ Georgia Tech \And
Bernardo Avila-Pires\\ DeepMind \And
Lindsey Kitchell\\ JHU-APL \And
Keith B. Hengen\\ WashU-St Louis \And
William Gray-Roncal\\ JHU-APL \And
Michal Valko\\ DeepMind \And
Eva L. Dyer\\ Georgia Tech
}

\begin{document}

\maketitle

\begin{abstract}
State-of-the-art methods for self-supervised learning (SSL) build representations by maximizing the similarity between different transformed “views” of a sample. Without sufficient diversity in the transformations used to create views, however, it can be difficult to overcome nuisance variables in the data and build rich representations. 
This motivates the use of the dataset itself to find similar, yet distinct, samples to serve as views for one another.
In this paper, we introduce Mine Your Own vieW (\alg), a new approach for self-supervised  learning that looks within the dataset to define diverse targets for prediction.
The idea behind our approach is to actively {\em mine views}, finding samples that are neighbors in the representation space of the network, and then predict, from one sample's latent representation, the representation of a nearby sample. After showing the promise of \alg on benchmarks used in computer vision, we highlight the power of this idea in a novel application in neuroscience where SSL has yet to be applied. 
When tested on multi-unit neural recordings, we find that \alg outperforms other self-supervised approaches in all examples (in some cases by more than 10\%), and often surpasses the supervised baseline. 
With \alg, we show that it is possible to harness the diversity of the data to build rich views and leverage self-supervision in new domains where augmentations are limited or unknown. 

\vspace{-2mm}
\end{abstract}
\vspace{-4mm}
\vspace{0.12in}

\section{Introduction}

\begin{figure*}[t!]
\centering
   \includegraphics[width=0.97\textwidth]{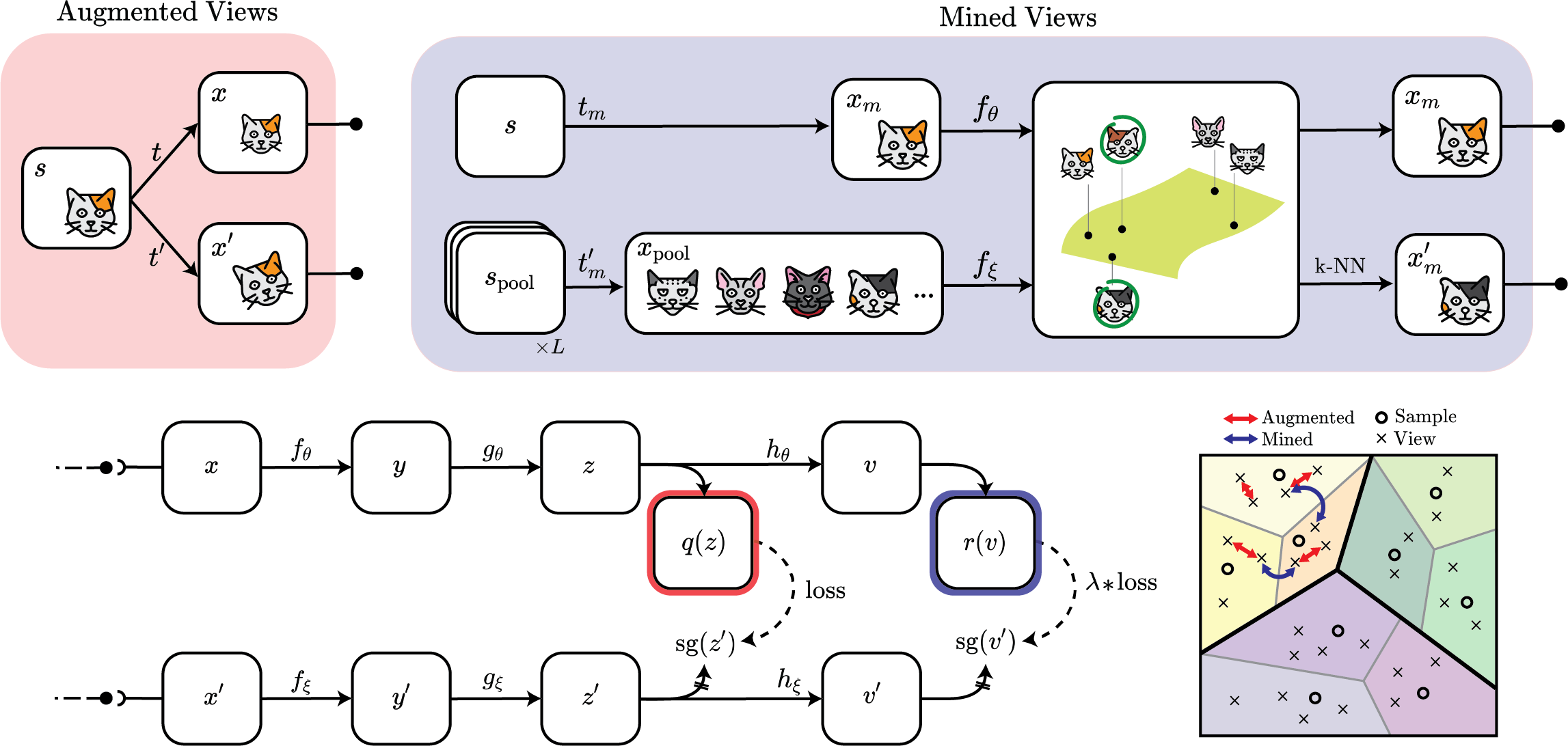}
   \caption{\footnotesize{\em Overview of our approach}. The architecture of the system, shown in the bottom row, consists of two networks, the online network (top) and the target network (below).
   There exists two sources of views, the augmented views block (top, red) and the mined views block (top, blue). Each type of view is handled by a dedicated predictor of the corresponding color. During mining, mined views are found by computing the $k$-nearest neighbors of the anchor online representation among the target representations of the pool of candidates. One of the nearest neighbors is randomly selected to be the mined view. On the bottom right, we illustrate the idea behind across-sample prediction and show how the two spaces emphasize different levels of similarity between data points. \label{fig:overview}}
   \vspace{-3mm}
\end{figure*}

Self-supervised learning (SSL) methods have made impressive advances on a wide range of tasks in vision \cite{doersch2015unsupervised,perez2017effectiveness,he2020momentum, chen2020simple, grill2020bootstrap, cai2020joint, song2020multi}, speech \cite{oord2018representation}, graphs \cite{velivckovic2018deep,zhu2020deep}, and reinforcement learning \cite{oord2018representation,guo2020bootstrap, schwarzer2020data}. 
This has been due, in part, to the simple paradigm of instance-level learning, where a representation is learned by maximizing the similarity between different transformed ``views''  of the same sample (positive examples). Contrastive learning methods compare positive examples to views of other samples (negative examples) and encourage them to have dissimilar representations \cite{he2020momentum, chen2020simple, oord2018representation, gutmann2010noise}, while more recent methods like \byol \cite{grill2020bootstrap}, \wmse \cite{ermolov2020whitening}, and \barlow \cite{zbontar2021barlow} show how this instance-specific approach can be implemented without the need for negative examples.

Augmentations are a key component of self-supervised methods; they establish the invariances learned by the network and control the richness of the learned representation. Thus, there are many cases where it is useful to {\em go beyond simple augmentations} to integrate more diverse views into learning \cite{wang2021stronger,kong2020flag}. At the same time, it can be challenging to find the right balance between augmentations that both introduce sufficient diversity and preserve the semantics of the original data. This is particularly true in new domains, like brain decoding, where we do not have prior knowledge to guide our search. Here, we ask whether diverse views can be found by {\em looking within the dataset}. Intuitively, other examples drawn from the dataset have the potential to satisfy both criteria: They offer more diversity, and when chosen correctly, they will share semantic content with the target sample.

In this paper, we introduce \textbf{M}ine \textbf{Y}our \textbf{O}wn vie\textbf{W} (\alg), a  self-supervised approach for representation learning that looks within the dataset for different samples to use as positive examples for one another. The idea behind our strategy is to {\em mine views}, or adaptively select other samples that are nearby in the latent space, and then use these mined views as targets for self-supervision.
To integrate both mined and augmented views into a unified framework,  we introduce a novel, cascaded dual projector architecture that learns to predict across augmented views of the {\em same sample} in the first part of the network, and then to predict across mined views of {\em different samples} through a separate projector/predictor that draws from the first projector's output (Figure~\ref{fig:overview}). 

To first test the method in domains where effective augmentations are well-established, we apply our approach to computer vision benchmarks, including CIFAR-10, CIFAR-100 and Tiny Imagenet. On these  benchmark datasets, we show that \alg is competitive with state-of-the-art methods like \byol \cite{grill2020bootstrap} and \simclr \cite{chen2020simple} and in many cases, outperforms these methods. 
After validating our approach in the image domain, we then turn our attention to {\em brain decoding from  multi-neuron recordings}, a novel application of SSL where diverse augmentations are unknown. We apply our approach to neural activities from the brains of non-human primates and rodents, where we show significant improvements over other approaches on two distinct brain decoding tasks (i.e., movement prediction from motor cortex, and sleep stage prediction from V1 and hippocampus). 
These results suggest that nearest-neighbor prediction can be a very effective tool for self-supervision in new domains where strong augmentations are not already established. 

Overall, we make the following contributions:
\vspace{-2mm}
\begin{itemize}
\item In Section~\ref{sec:approach}, we introduce \alg, a new approach for adaptively finding views from distinct samples in the dataset and using them as positive examples for one another. We introduce a novel cascaded dual projector architecture that builds on \byol to integrate augmented and mined views without the need for negative examples.

\item After validating our approach on standard datasets used in computer vision, in Section~\ref{sec:neuraldata}, we show how SSL and \alg can be applied to multi-neuron recordings. To the best of our knowledge, this is the first time that SSL has been applied to these types of brain  datasets that capture activity at the level of individual neurons. We establish a set of universal augmentations, that can be successfully applied to datasets spanning non-human primate, rat, and mouse.

\item In our experiments on neural datasets (Section~\ref{sec:neuralresults}), we show that by linking ``semantically close'' yet temporally separated brain states, \alg yields significant improvement 
in the decoding of behavior when compared to other self-supervised approaches. We also observe that in some datasets, the linear readouts from our representation layer provide better decoding performance than supervised methods, suggesting that \alg can be a powerful tool for reading out information from neural circuits.
\end{itemize}


\section{Mine Your Own vieW (\alg) }
\label{sec:approach}
\vspace{-1mm}
In this section, we introduce \alg, our proposed self-supervised approach for across-sample prediction (see Figure~\ref{fig:overview}). 
A PyTorch implementation of \alg is made available at:
\url{https://nerdslab.github.io/myow/}.

\vspace{-1mm}
\subsection{Combining augmented and mined views through cascaded predictors}
\vspace{-2mm}

To build a representation, we will leverage the predictive framework introduced in \byol \cite{grill2020bootstrap} which aims to maximize similarity across {\em augmented views}.
Instead of relying solely on instance-level augmentations, \alg finds {\em mined views}, or views of different samples that are close in the latent space.
We now provide a detailed overview of our method starting with the architecture, and then describing our view mining approach (see Appendix~\ref{app:alg} for pseudocode).

{\bf View generation.} Given a sample ${\bf s} \in \mathcal{D}$ from our dataset, we generate two augmented views $\x, \x'$ using transformations $t, t'$ sampled from a set $\mathcal{T}$.
A third view $\x_m$ of the same example is also generated, while the mined view $\x_m'$ is of a different sample ${\bf s}'$ {\em selected} from the dataset. The transformations $t_m, t_m'$ to produce these views are sampled from a set $\mathcal{T}_m$ which is not necessarily the same as $\mathcal{T}$.
Different heuristics can be designed to mine views; in the next section, we present a simple nearest neighbor strategy, which uses points that are nearby in the representation space of the network to serve as positive examples for each other.

{\bf Dual deep architecture.} Both types of views are fed through {\em online} and {\em target} networks, parameterized by weights $\theta$ and $\xi$, respectively. The encoders produce representations $\y = f_\theta( \x)$ and $\y' = f_\xi( \x')$, which are then passed through a projector to obtain $\z = g_\theta(\y)$ and $\z' = g_{\xi} (\y')$. Mined views are further projected in secondary spaces to obtain ${\bf v}_m = h_\theta(\z_m)$ and ${\bf v}_m' = h_\xi(\z_m')$.
The projections in the target network act as  targets for their respective predictors: $q_\theta$ forms predictions across augmented views and $r_\theta$ forms predictions across mined views.

{\bf Loss function.} \alg learns a representation by minimizing both augmented and mined prediction errors through the following loss:
\begin{equation}
\begin{split}
    \mathcal{L} =\ 
     \underbrace{d(q_\theta(\z), \z')}_{\text{Augmentation Loss}}
    + \lambda~  \underbrace{d( r_\theta\left( {\bf v}_m \right ),   {\bf v}_m^\prime)}_{\text{Mining Loss}}, \quad \text{with} \ \  d({\bf u},\boldsymbol{\nu}) = - \frac{\left\langle {\bf u},\boldsymbol{\nu}\right\rangle}{\left\|{\bf u}\right\|_{2}\left\|\boldsymbol{\nu}\right\|_{2}},
    \label{eq:combinedloss}
\end{split}
\end{equation}
where $\lambda$ is a weight that regulates the contribution of the mined views in the objective; in practice, $\lambda$ has an initial linear warmup period of a few epochs. Just as in \byol, we symmetrize the distance between augmented views by feeding $\x'$ and $\x$ to the online and target network, respectively.

We use the same approach for optimizing the online and target networks as proposed in \byol. The loss $\mathcal{L}$ is optimized only in terms of $\theta$ and $\xi$ is updated according to a moving average of $\theta$. In particular, we update the online and target networks according to the following:
\begin{equation}
\label{eq:update}
\theta \leftarrow \textrm{optimize}(\theta, \nabla_{\theta} \mathcal{L}, \eta),~~
\xi \leftarrow \tau \xi + (1-\tau) \theta,
\end{equation}
where  $\tau\in [0,1]$ is a momentum parameter, and $\eta$ is the learning rate used to optimize the weights of the online network. We point the reader to a discussion of the cost and benefits of different components of this dual network implementation (i.e., stop gradient, predictor, momentum) \cite{chen2020exploring}.

\vspace{-1mm}
\subsection{How to mine views}
\vspace{-2mm}

{\bf Randomized nearest-neighbor selection approach.} 
\alg adaptively ``mines'' samples in the dataset that are neighbors in the representation space and uses them as positive examples. 
One could imagine many strategies for doing this; we show that a simple random k-nearest neighbor (k-NN) strategy suffices.
Specifically, given an anchor sample ${\bf s}$, we draw a set of $L$ candidate samples and apply transformations sampled from a set $\mathcal{T}_m$.\footnote{In general, the set of transformations that we use for mined views, $\mathcal{T}_m$, can be different from the set $\mathcal{T}$ used for augmented views; particularly in the case of images, we empirically find that the use of simpler transformations is more favorable.} The anchor sample is passed through the online encoder to obtain its representation $\y_m = f_\theta(\x_m)$, where $\x_m=t_m({\bf s})$ and $t_m \sim \mathcal{T}_m$. The candidate views $\{ \x_{j} \}$ (generated from other samples) are projected in the target encoder's space to obtain $\mathcal{S} = \{ f_\xi(\x_{j})\}_{L}$. The $k$-nearest neighbors of the anchor representation $\y_m$ are computed from this set $\mathcal{S}$ and one of these neighbors is randomly selected as the mined view $\x_m^{\prime}$. 

{\bf Controlling stochasticity in mining.}~
There are two main parameters that must be specified for mining, the number of nearest neighbors ($k$) and the number of samples that are considered as candidates for mining ($L$). Both of these parameters control the diversity and randomness of which views may be selected. Only a fraction of the dataset ($L/N$) is used during the mining process, the smaller this fraction gets, the more stochastic the mining becomes: at the end of training, each sample would have seen a large and diverse set of mined views. In the case of the image datasets we study, we are able to use a pool of candidates of size equal to the batch size $L=B=512$ with $k=1$. 
On neural datasets, we find that slightly higher values of $k$ are more favorable, suggesting that more stochasticity is helpful in this case. In all of our experiments, we find that \alg can be effectively trained using $L=B$.

{\bf Defining which samples can be selected through mining.}
When mining views, our algorithm can flexibly accommodate different constraints into our mining procedure. While not necessary in images, when mining in temporal data (like our neural examples), we know that temporally close data points can be selected as augmentations and thus it is useful to restrict the mining candidates to samples that are either farther in time from the anchor sample or in entirely different temporal sequences. Further details on our mining procedure can be found in Appendix~\ref{app:mining-graph}; we note that the same global pool of candidates of size $L$ is used for all samples in a batch.

 \vspace{-1mm}
\subsection{Memory and computational requirements}
 \vspace{-2mm}
In our experiments, the pool of candidates is resampled on-the-fly at each iteration and thus \alg does not require a memory bank. While there is an additional, but negligible (less than 2\%), memory overhead due to the k-NN operation, the memory requirements for training \alg are not different from \byol’s when $L \leq B$. This is because augmented and mined views are forwarded sequentially through the network and gradients are accumulated before updating the weights. To reduce the extra computational overhead due to mining, we use the candidates' target representations instead of their online representations and avoid an extra forward pass. We empirically find that mining in either the online or target network leads to similar results (Appendix~\ref{app:minequality}) and thus use this strategy in practice. In this case, \alg  requires 1.5x computation time when compared to \byol. When memory is not an issue, computation time can be reduced significantly by feeding in all views at the same time. When using a multi-GPU setup, we distribute the computation of the candidate’s representations over all GPUs and then have them broadcast their local pools to each other, effectively building a pool of mining candidates of larger size.

\section{Evaluations}

In order to evaluate our approach, we first test it on benchmark datasets used for image recognition in computer vision. After we establish the promise of our approach on images, we then focus our attention on a novel application of SSL to  decoding latent variables from neural population activity. 


 \vspace{-1mm}
\subsection{Image datasets: Comparisons and ablations}
\label{sec:images}
 \vspace{-2mm}

{\bf Experimental setup.}~
To train our model and other SSL approaches on natural images, we follow the procedures reported in previous work \cite{chen2020simple, chen2020exploring, huang2021selfadaptive}, both to augment the data and evaluate our models (see Appendix~\ref{app:exp-images}).
We train the networks for $800$ epochs and use a batch size of 512. When mining, we use an equally sized pool of candidates $L=512$, as well as $k=1$ and $\lambda=0.1$. During training we use an SGD optimizer with a learning rate of $\eta=0.03$ to update the online network, and a moving average momentum of $\tau=0.996$ for the target network.
For all ResNet-18 and ResNet-50 experiments, we train using 1 and 2 GTX 2080Ti GPU(s), respectively. We assess the quality of the representations by following the standard linear evaluation protocol: a linear layer is trained on top of the frozen representation, and the accuracy is reported on the validation set. Models trained on CIFAR-100 are also evaluated on CIFAR-20 which aggregates labels into 20 superclasses. 

{\bf Results on natural images.}~ In our experiments, we compare \alg with both \byol, and \simclr on CIFAR-10, CIFAR-100 and Tiny ImageNet (Table~\ref{tab:imageres}). Consistently, \alg yields competitive results with these state-of-the-art methods, and  outperforms \byol even when they share the same random seed and the same hyper-parameters. We rule out the possibility that \alg simply benefits from an effectively higher batch size by conducting experiments where the batch size or number of epochs used in \byol is increased by $50\%$ (Appendix~\ref{app:compute}). More significantly, we find, for the CIFAR-10 experiment, that \alg surpasses \byol's final accuracy only after $300$ epochs, which, in this case, largely justifies the additional computational cost of our approach.
When we consider a limited augmentation regime (Table~\ref{tab:imagecrop}), we find that \alg increases its gap above \byol. Overall, we find that \alg  provides competitive performance on the vision datasets we tested.

\begin{figure}[t!]
\centering
\vspace{-1mm}
    \includegraphics[width=0.93\textwidth]{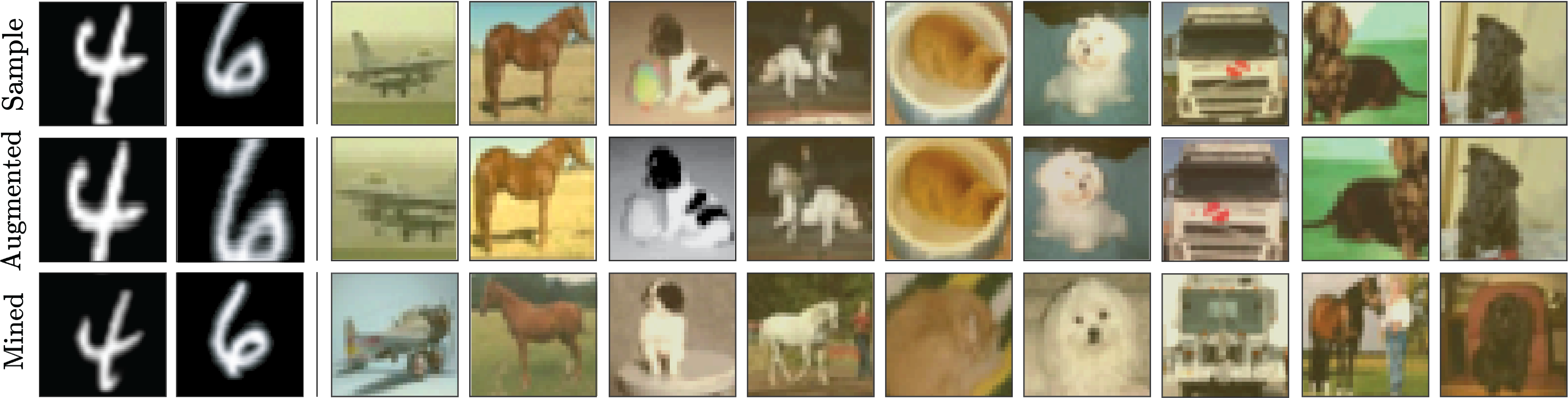}
    \caption{\footnotesize {\em Examples of mined views from MNIST (left) and CIFAR-10 (right).}}
    \label{fig:imageres}
    
\vspace{-0.1in}
\end{figure}

{\bf Examining mined views.}~Figure~\ref{fig:imageres} highlights examples of views mined during training, where we can see the rich semantic content shared within each pair. Even when mined views are not from the same class, we find other semantic similarities shared between the views (see the penultimate column where we select a Dachshund dog and the horse with similar body shape and color through mining). 
While we do find that the mining process does not always select positive examples from the same class (refer to Appendix~\ref{app:minequality}), the presence of these across-class predictions does not seem to hinder performance.

{\bf Ablations.}~Our architecture integrates mined views through a second cascaded projector/predictor. On both MNIST and CIFAR-10, we performed architecture ablations to study the role of our cascaded architecture compared to a single projector or parallel dual projectors (Appendix~\ref{app:projector}).
Our experiments reveal that all three configurations (cascaded, single, parallel) lead to an improvement over the \byol baseline in CIFAR-10, with the cascaded architecture showing the best performance. We also perform ablations on the class of transformations $\mathcal{T}_m$ used for mined views (Appendix~\ref{app:minequality}), and find that, when training on the CIFAR-10 dataset, the use of minimal to no transformations yields the best result.


\begin{table}[t!]
\centering
\caption{\footnotesize{\em  Accuracy (in \%) for classification on CIFAR-10, CIFAR-100 and Tiny Imagenet.} We report the linear evaluation accuracies for different architectures and datasets. For CIFAR-100, we report both accuracies under linear evaluation on CIFAR-100 and CIFAR-20. Results for \simclr are reported from \cite{ermolov2020whitening}.}
\vspace{0.1in}
\resizebox{\textwidth}{!}{
\begin{tabular}{l|cccc|ccc}
\hline
       & \multicolumn{4}{c|}{\em ResNet-18}      & \multicolumn{3}{c}{\em ResNet-50}   \rule{0pt}{2ex}  \\ 
Method & CIFAR-10       & CIFAR-100      & CIFAR-20       & Tiny ImageNet  & CIFAR-10       & CIFAR-100      & CIFAR-20  \\ \hline
\simclr * & 91.80          & 66.83          & -              & 48.84          & 91.73          & -              & -           \rule{0pt}{2ex}   \\
\byol   & 91.71          & 66.70          & 76.90          & 51.56          & 92.12          & 67.87          & 77.38          \\
\alg    & \textbf{92.10} & \textbf{67.91} & \textbf{78.10} & \textbf{52.58} & \textbf{93.18} & \textbf{68.69} & \textbf{78.87} \\ \hline
\end{tabular}

}
     \label{tab:imageres}
\end{table}


\begin{table}[t!]
\vspace{-0.1in}
\caption{\footnotesize{\em  Accuracy (in \%) for different classes of transformations.} We report the linear evaluation accuracies for \byol and \alg trained on CIFAR-10 using ResNet-18.}
     \label{tab:imagecrop}
\centering
\vspace{0.1in}
\resizebox{0.65\textwidth}{!}{
\begin{tabular}{lcccc}
\hline
       & Original & Remove Grayscale & Remove Color & Crop only  \rule{0pt}{2ex}  \\ \hline
\byol  & 91.71    & 88.04            & 87.13        & 82.10  \rule{0pt}{2ex}    \\
\alg   & 92.10    & 89.16            & 89.38        & 84.82     \\ \hline
\end{tabular}}

\end{table}


 \vspace{-1mm}
\subsection{Neural datasets: Identifying  classes of augmentations}
\label{sec:neuraldata}

After establishing our method on image datasets, we set out to test our approach on multi-neuron recordings.  
As this is the first attempt at leveraging a self-supervised learning framework for neural data of this nature, our first goal was to establish simple yet general classes of augmentations that can be utilized in this application.

{\bf Neural datasets and decoding tasks.}~
In our experiments, we consider a total of six neural datasets from both non-human primates and rodents.

\begin{figure}[t!]
    \centering
    \includegraphics[width=\textwidth]{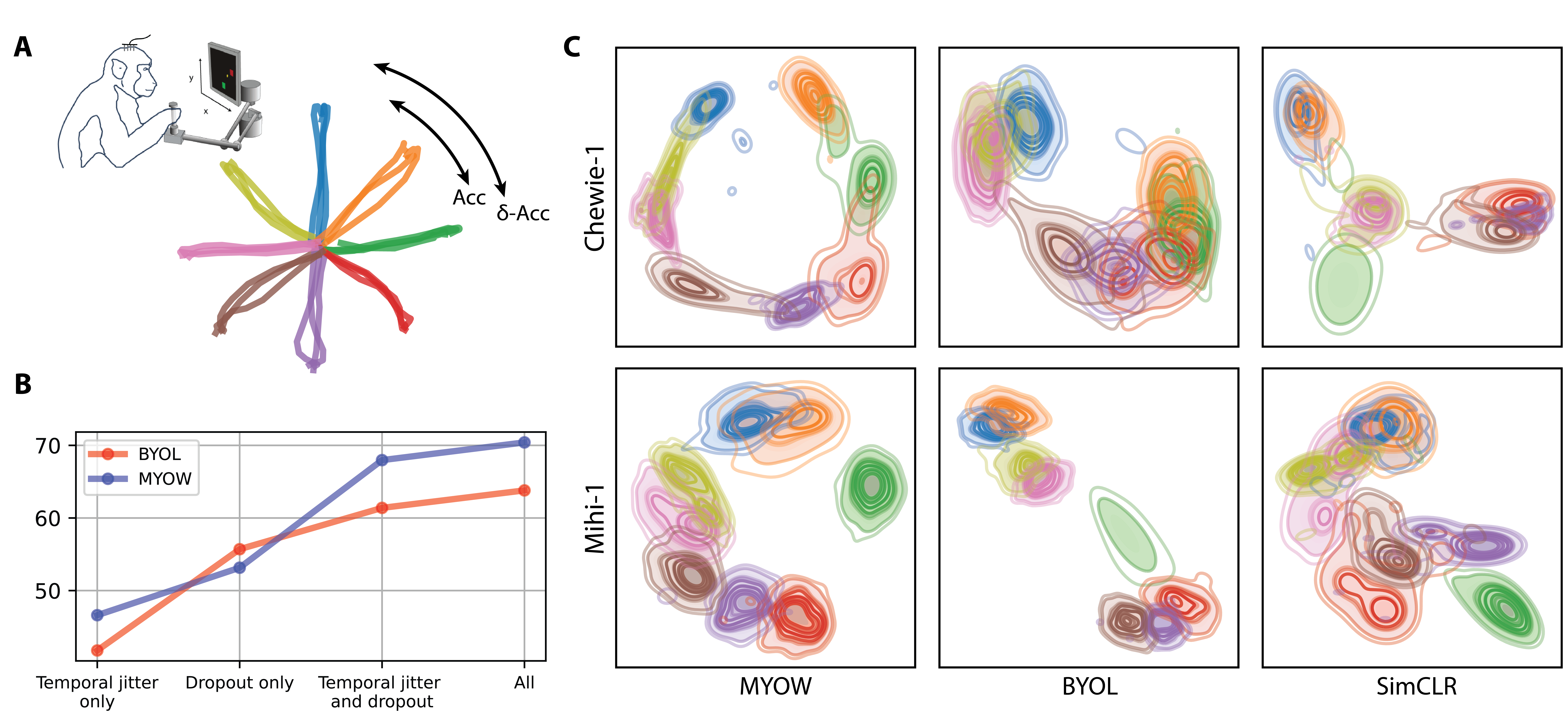}\vspace{-1mm}
    \caption{\label{fig:monkey}\footnotesize {\em } (A) Sketch of a primate performing a reach task with sample joystick trajectories depicting the center-out reach movement. (B) Increase in decoding accuracy of \byol and \alg as we progressively introduce new augmentations. (C) Visualization of the learned representations obtained for the two primates Chewie and Mihi when different SSL methods are applied (embeddings are obtained via t-SNE). This shows how \alg reveals the underlying structure of the task, as clusters are organized in a circle. \label{fig:monkeyres}}
\end{figure}

{\em 1) Reaching datasets.}~The first datasets we will consider are acquired from the primary motor cortex (M1) of two non-human primates (NHPs), Chewie and Mihi, while they repeatedly made reaching movements towards one of eight targets \cite{dyer2017cryptography}. We call each repetition a ``trial''. Spiking activity of $d$ single neurons is recorded during each reach trial, Figure~\ref{fig:monkey}-A shows some instances of the trajectory of the joystick during movement. Data was collected at two different dates (77 days apart for Chewie, 3 days apart for Mihi), resulting in two datasets per primate, each targeting a different set and number of neurons in the same part of M1. The activity of neurons was spike-sorted and binned into 100ms intervals to generate around $1.3$k $d$-dimensional vectors per dataset. 
To measure representation quality, we will define our target downstream task as the {\em decoding of the intended reach direction} from the neural activity during the movement.

{\em 2) Sleep datasets.}~The second datasets that we will consider are collected from rodent visual cortex (V1) and hippocampus (CA1) during free behavior over 12 hours \cite{HENGEN2019}. Here, neural activity was binned into 4s intervals to produce firing rates for 42 and 120 single neurons, for a rat and mouse, respectively.
To measure the quality of representations learned, we will define our downstream task as the {\em decoding of the arousal state} of the rodent into one of three classes: rapid eye movement (REM) sleep, non-REM sleep, or wake \cite{HENGEN2016, HENGEN2019}. 

{\bf Experimental setup.}~ For all datasets, we use multi-layer perceptrons (MLPs) as encoders with a representation size of 64 and 32, for primate and rodent data respectively. We train the networks for $1000$ epochs and use a batch size of 512. When mining we use an equally sized pool of candidates $L=512$, as well as $k=3$ and $\lambda=0.1$.
During training we update the online network using AdamW with a learning rate of $\eta=0.02$ for primates and $\eta=0.001$ for rodents and weight decay of $2 * 10^{-5}$, and use a moving average momentum of $\tau=0.98$ for the target network. Each dataset is temporally split into (70/10/20\%) train/validation/test sets.
More details on the datasets and experimental setup can be found in Appendix~\ref{app:exp-neural}.

{\bf Augmentations for spiking neural data.}~
While self-supervised approaches have not been applied to the multi-neuron recordings that we consider, we take cues from other domains (video, graphs), as well as previous work on electroencephalogram (EEG) data \cite{cheng2020subject, sarkar2020ecg}, to define simple classes of augmentations for our datasets. Specifically, we consider four different  types of augmentations: (i) {\em Temporal Jitter}--stochastic backward or forward prediction of nearby samples within a small window around the sample, (ii) {\em Dropout}--masking neurons with some probability, and (iii) {\em Pepper}--sparse additive noise, and (iv) {\em Noise}--additive Gaussian noise. 

We test the inclusion and combination of these different augmentations, first on our \byol backbone which uses augmented views only (Figure~\ref{fig:monkeyres}-B). While we find that temporal jitter alone is insufficient to drive learning, when we combine both jitter and dropout,  we  see a substantial increase in decoding accuracy and qualitative improvements in the resulting representations. In this case, our baseline SSL method, \byol, quickly starts to create meaningful predictive relationships between data, as evidenced by our decoding results and qualitative evaluations of the representations (Appendix~\ref{app:augmentations}). 
As we include additional augmentations ({\em Noise} + {\em Pepper}), the performance increases further, but by smaller margins than before. In general, we see these same trends observed throughout our remaining primate datasets and in our experiments on rodent (see Appendix \ref{app:augmentations}), suggesting that these classes of transformations are good candidates for building SSL frameworks for neural activity.

After establishing a good set of simple augmentations, we then integrate {\em mined views} with \alg (Figure~\ref{fig:monkey}-B, blue). In this case, we can interpret mined views as {\em nonlocal brain states} that are not temporally close but can be semantically similar. For instance, in our reaching datasets, \alg will mine outside of the current reach and look for other samples that it can use to build a more unified picture of the brain states as they evolve. Through combining simple augmentations with nonlocal samples with \alg, we provide an impressive boost in performance over \byol on this application. 


\begin{table}[h!]
        \centering
         \caption{\footnotesize{\em Accuracy (in \%) in the prediction of reach direction from spiking neural activity.}~ \label{tab:monkeyres}}
         \vspace{0.1in}
        \label{tab:monkey}
        \resizebox{0.8\textwidth}{!}{
        \begin{tabular}{lcccccccccc}
        \hline
        & \multicolumn{2}{c}{Chewie-1}          & \multicolumn{2}{c}{Chewie-2} & \multicolumn{2}{c}{Mihi-1} & \multicolumn{2}{c}{Mihi-2}  \rule{0pt}{2ex}    \\
                        & Acc & \deltacc & Acc & \deltacc & Acc  & \deltacc & Acc & \deltacc\\ \hline
         \texttt{Supervised}      & 63.29  & 77.22  &  72.29  &  81.51 & 63.64  & 79.02  & 61.49 & 68.44 \rule{0pt}{2ex}    \\
        \texttt{pi-VAE}      & 65.63 & {\bf 82.62} & 60.60 & 74.64 & 62.44 & 77.12 & 63.26 & 77.58 \\
        \hline
        \texttt{AE}     & 48.40   & 67.51   &  46.79 & 65.84  &  50.94  &  68.03 & 55.19 &  74.98 \rule{0pt}{2ex}    \\
        \texttt{RP}          & 59.21 & 78.10  & 50.69  & 60.01  & 57.78  & 76.03  & 53.76 & 71.34\\
         \texttt{TS}          & 60.16 & 78.76  & 49.48  & 63.55  & 59.23  & 76.98  & 54.10 & 71.65\\
        \simclr          & 61.36 & 79.40  & 51.62  & 65.01  & 59.41  & 77.82  & 56.29 & 74.57\\
        \byol           & 66.65  & 78.17  & 64.56  & 77.22  & 72.64  & 85.14  & 67.44 & {\bf 82.17} \\
        \alg            &  {\bf 70.54}  &  79.99  & {\bf 72.33}  &  {\bf 84.81}  &  {\bf 73.40}  &  {\bf 85.58}  &  {\bf 71.80} & 81.96 \\ \hline
        \end{tabular}}
        
\end{table}

 \vspace{-1mm}
\subsection{Neural datasets: Examining and comparing representation quality}
\label{sec:neuralresults}
 \vspace{-2mm}

Next, we will test the representational quality of our model by asking how well relevant behavioral variables can be linearly decoded from the learned representations.

{\bf Decoding movements from the primate brain.}~
In the reaching datasets that we consider here, there is a direct connection between the neural state (brain activity across many neurons) and the underlying movements (behavior). Thus, we wanted to assess the quality of the representations learned from these datasets by asking how well we can predict the reach direction from neural activity. If we have a good representation, we should be able to better separate reach direction from the neural activities. To quantify this, we will use a linear readout to predict the cosine and sine of the reach direction, and report the classification accuracy. We also introduce a slightly relaxed accuracy metric that we call the \deltacc~(akin to Top-k), which has a larger true positive acceptance range, as can be seen in Figure~\ref{fig:monkey}-A. (see Appendix~\ref{app:exp-neural} for a formal definition).

We compare our approach with several self-supervised methods, including state-of-the-art methods \byol and \simclr,
as well as two widely used self-supervised tasks recently applied to EEG data called Relative Positioning (\texttt{RP}) and Temporal Shuffling (\texttt{TS}) \cite{banville2019eeg}. \texttt{RP} trains the network by classifying whether two samples are temporally close, while \texttt{TS} takes in three samples and learns whether they are in the right order or if they are shuffled. In addition to these self-supervised methods, we also train a Multi-layer Perceptron (MLP) classifier (\texttt{Supervised}) using weight regularization and  dropout (in nodes in intermediate layers in the network), an autoencoder (\texttt{AE}), and a state-of-the-art  supervised approach for generative modeling of neural activity (\texttt{pi-VAE}) that leverages behavioral labels to condition and decompose the latent space \cite{zhou2020learning}.

We find that \alg  consistently outperforms other approaches and that contrastive methods that rely on negative examples (\texttt{SimCLR},  \texttt{RP} and  \texttt{TS}) fall behind both \alg and \byol. We also find that \alg generalizes to unseen data more readily than others; in some cases, beating supervised approaches by a significant margin, with over 10\% on both Mihi datasets. 
When we consider \deltacc, our method scores above 80\% on all datasets, outperforming the supervised baseline by over 10\% on Mihi-2. 
These results are even more impressive considering that we only tune augmentations and hyperparameters on Chewie-1 and find that \alg consistently generalizes across time and individuals. We thus show that by integrating diverse views (across trials) through mining into our prediction task, we can more accurately decode movement variables than supervised decoders.

When we visualize the learned representation in Figure~\ref{fig:monkeyres}-C, we notice that \alg organizes representations in a way that is more reflective of the global task structure, placing reach directions in their correct circular order. In contrast, we find that in both individuals, other methods tend to distort the underlying latent structure of the behavior when visualized in low-dimensions (Appendix~\ref{app:latent-neural}). 
We conjecture that across-sample predictions (including those across different reach directions), may be responsible for capturing this kind of higher-level structure in the data.

\begin{figure}[t!]
  \begin{minipage}[h]{0.59\textwidth}
    \centering
    \includegraphics[width=\textwidth]{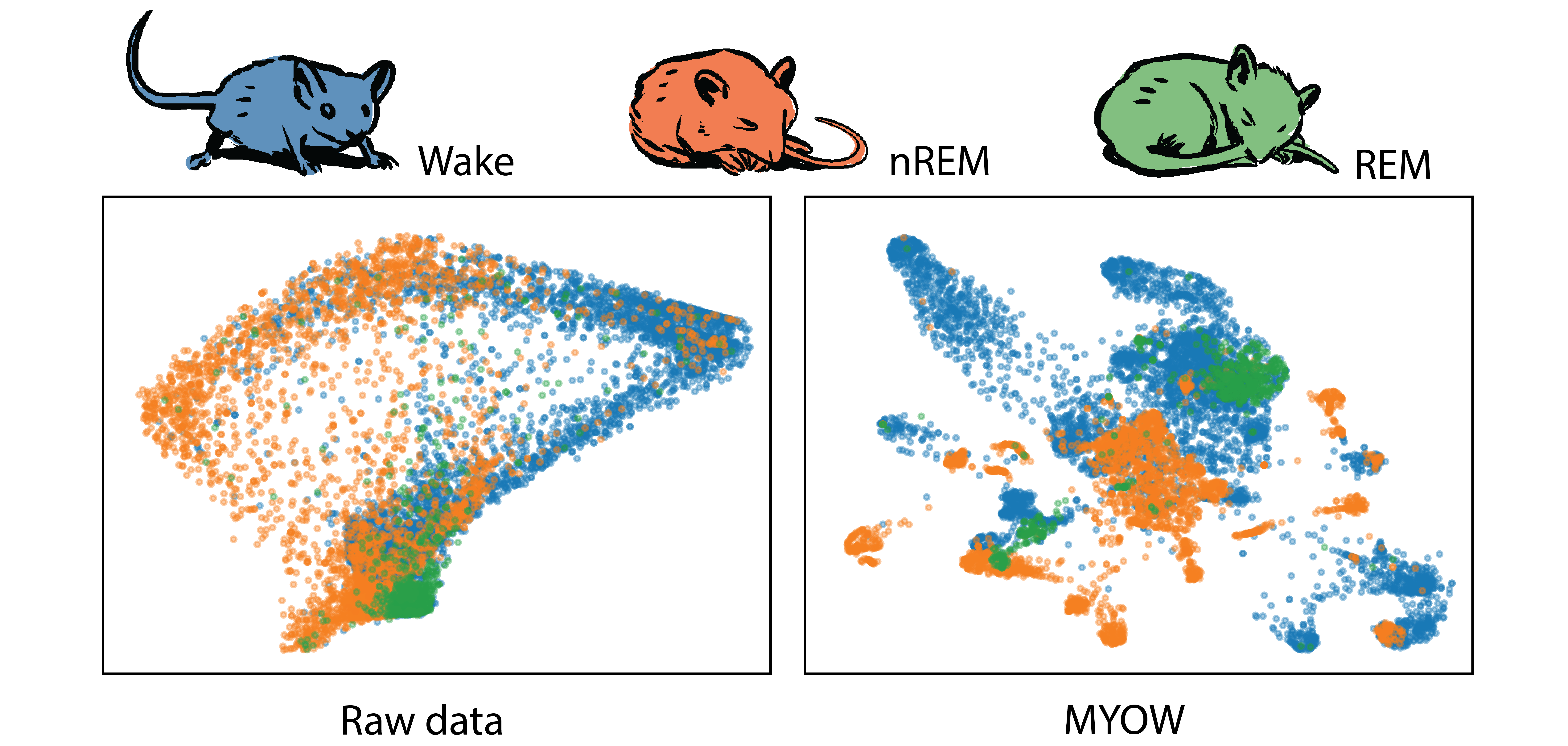}
  \end{minipage}
  \hfill
  \begin{minipage}[h]{0.36\textwidth}
    \centering
        \resizebox{1.0\textwidth}{!}{\begin{tabular}{lcc}
        \hline
         & Rat-V1 & Mouse-CA1\\\hline
        \texttt{Supervised}  & 86.34 & 93.01\\
        \texttt{pi-VAE} & 73.10 & 82.48\\\hline
        \texttt{AE}    & 34.17 & 57.73\\
        \texttt{RP}  & 82.93 & 82.12 \\
        \texttt{TS}&  82.45 & 81.93\\
        \simclr  & 81.03 & 81.94 \\
        \byol  & 85.42 & 93.24\\
        \alg & {\bf 88.01} & {\bf 93.70}\\ \hline
        \end{tabular}}
    \end{minipage}
    \vspace{-1mm}
  \caption{\footnotesize{ (Left) Visualizations of raw and \alg latent spaces (using t-SNE) of 12 hour recordings from mouse (CA1) during free behavior, including sleep and wake. One variable of interest is the arousal state (REM, nREM, Wake). (Right) F1-score (in \%) in the prediction of arousal state from spiking neural activity.\label{fig:mouse}\label{tab:res} }}  
\end{figure}

{\bf Decoding arousal states from the rodent brain during free behavior.}~
Next, we applied \alg to datasets from the rodent cortex and hippocampus, where we test our ability to decode arousal states (REM, nREM, Wake) from the learned representations. Despite the strong class imbalance, the trends are similar to that of our earlier experiments, with \alg providing robust performance, exceeding that of the supervised baseline, and outperforming other self-supervised methods. 

In these datasets, the animal is ``untethered'' and can roam around in its cage without any task or  explicit instructions. In these free-behaving conditions, we find a great deal of variability in the latent state beyond the coarse labels that we have access to. When we visualize the representation learned by \alg in Figure~\ref{fig:mouse}, we find that the network separates different parts of the behavior space, revealing subspaces of neural states that are otherwise unobservable when examining the embeddings of the raw data. 


\section{Related Work}\label{sec:background}

{\bf Self-supervised learning.}~SSL aims to learn representations of unlabeled data that are useful for downstream tasks. While early work utilized proxy tasks for self-supervision \cite{gidaris2018unsupervised,doersch2016unsupervised}, instance discrimination-based SSL methods \cite{chen2020simple, he2020momentum, grill2020bootstrap, zbontar2021barlow} have emerged as the state-of-the-art for representation learning, showing tremendous success and moving towards closing the gap with supervised learning. Conceptually, these approaches treat each instance in the dataset as its own class. A given sample is transformed to create distinct {\em positive views}, which are encouraged to be close in terms of their representations, while negative pairs are pushed apart. 
\byol \cite{grill2020bootstrap}, \texttt{SimSiam} \cite{chen2020exploring}, and more recently \barlow \cite{zbontar2021barlow} move away from the explicit contrastive framework and the reliance on negative samples by employing different strategies that avoid collapse in the representation.
The precise mechanisms underlying the success of \byol \cite{grill2020bootstrap} are still unclear and have been the subject of recent theoretical and empirical studies \cite{tian2021understanding, richemond2020byol}.

{\bf Connections to mining hard negatives in contrastive learning.}~In contrastive learning, it is a commonly held belief that the use of large numbers of negative examples is necessary to introduce enough ``hard negative examples'' into learning. Thus, there has been interest in nearest-neighbor sampling and mixing to define {\em hard negative examples} \cite{zhang2018mixup,kalantidis2020hard, robinson2021contrastive} instead of just relying on larger batch sizes. Interestingly, the mined views in \alg can be considered as {\em harder positive examples}, but are different from their negative counterpart in that they define a new type of views. 

{\bf Clustering-based SSL and local aggregation (LA).}~
Clustering-based representation learning methods are different from instance-specific contrastive methods in that they do not compare pairs of samples directly, but do it through the use of prototypes or pseudolabels. DeepCluster \cite{caron2018deep}, for example, uses k-means assignments as pseudolabels for training. LA \cite{zhuang2019local} leverages neighbors to guide learning by defining two sets of neighbors, close and background neighbors, encouraging close neighbors to be nearby while pushing them away from a set of background neighbors. More recently, SwAv \cite{caron2020unsupervised} simultaneously learns a set of prototype vectors and enforces consistency between cluster assignments of two positive views.

Like many of these methods, we select samples with similar embeddings and use them to adaptively link data samples in the latent space. However, instead of using a small number of prototypes to cluster the representations, we use neighbors in the representation space as positive views for prediction and do not force any kind of explicit clustering. Moreover, because our model is built on \byol, {\em we do not require negative examples} and also avoid the introduction of more complex distance measures to establish contrast (e.g., close vs. background neighbors).

{\bf Applications of SSL in neuroscience and biosignal analysis.}~
Previous work in self-supervised and contrastive learning for sequential data often leverages a slowness assumption to use nearby samples as positive examples and farther samples as negative examples \cite{oord2018representation,sermanet2018timecontrastive,dwibedi2019learning,Le_Khac_2020,banville2020uncovering}. 
Contrastive predictive coding (\texttt{CPC}) \cite{oord2018representation} 
further leverages the temporal ordering in sequential data by building an autoregressive (AR)-model that predicts future points given previous observed timesteps.
In reinforcement learning, \texttt{PBL} \cite{guo2020bootstrap} also uses a similar strategy, however, they show similarly to \byol that negative examples are not needed to learn a good representation.

In \cite{banville2020uncovering}, the authors test different temporal contrastive methods ( \texttt{RP},  \texttt{TS} and \texttt{CPC}) on EEG datasets. 
They find that, despite the additional complexity afforded by TS and \texttt{CPC}, these approaches perform similarly to RP in their experiments on sleep decoding from the human brain. In \cite{cheng2020subject}, they propose a contrastive learning method for EEG that also leverages subject-level information to build representations.
Our approach shares similarity with these existing approaches in how we build augmented views for neural data. However, \alg goes beyond these temporally local predictions to incorporate nonlocal time points as positive examples. We show that non-local predictions across samples can be used to significantly boost performance for our neural datasets, and thus we expect that nearest-neighbor based approaches could also be used to extend these previous applications of SSL in neuroscience.


\section{Conclusion}
This paper introduces a new method for SSL that integrates diverse across-sample views into learning through a novel cascaded architecture. We show that our approach can be used to learn meaningful representations on a variety of image and neural datasets. 

This paper provides an important first step towards applying self-supervised methods to learn representations of neural activity. For these datasets, we establish general classes of augmentations and study the impact of these augmentations on diverse neural recordings.
Our results in this domain are compelling: we typically obtain better generalization than supervised methods trained with dropout and weight decay. Through the inclusion of temporal structure into our framework and architecture, we may be able to improve this approach even further and capture dynamics over longer timescales. 

In our application to spiking neural data, we demonstrate that both dropout and temporal augmentations are necessary for building meaningful representations of different brain states. Similarly in neural circuits, neurons are unable to send direct signals to every other neuron in a downstream population; thus, target areas receiving signals may need to predict future brain states from partial information \cite{rao1999predictive}. Our results suggest that it may be fruitful to try to understand how brains may leverage dropout to build predictive representations, and that a theoretical understanding of SSL might yield insight into these processes.

\section*{Acknowledgements}
This project was supported by NIH award 1R01EB029852-01, NSF award IIS-1755871 and IIS-2039741, as well as generous gifts from the Alfred Sloan Foundation and the McKnight Foundation.


\section{Acknowledgements}
ELD, MA, CHL, KBH, and KBN were supported by NIH-1R01EB029852. KBH and KBN were supported by NIH-1R01NS118442. This work was also supported by an award from the McKnight Foundation and Sloan Foundation. We would like to thank Bilal Piot for helpful suggestions on the manuscript.

\bibliographystyle{ieeetr}
\bibliography{0-main.bib}

\onecolumn
\newpage
\renewcommand{\thefigure}{S\arabic{figure}}
\setcounter{figure}{0}
\setcounter{table}{0}
\renewcommand{\thetable}{S\arabic{table}}%

\section*{\Large Appendix}
\appendix

\section{Algorithm}\label{app:alg}

\begin{algorithm}
\caption{\hspace{2mm} \texttt{Mine Your Own vieW} - \alg \label{alg:myow}} 

\Input{~\texttt{Dataset} ${\cal{D}}$;~ \texttt{online network} $f_{\theta},g_{\theta},h_{\theta}$;~ \texttt{target network} $f_{\xi}, {g}_{\xi}, h_{\xi}$;~ \texttt{dual predictors} $q_{\theta}, r_{\theta}$;~ \texttt{learning rate} $\eta$; \texttt{momentum} $\tau$;~ \texttt{mining weight} $\lambda$;~ \texttt{batch size} $B$;~ \texttt{pool batch size} $L$.}

\Init{$\xi\leftarrow \theta$}{}
\While{not converging}{
\Comment{\texttt{Augment views}}
\texttt{Fetch a mini-batch}  $\{{\s}_i\}_B$ \texttt{from} $\cal{D}$\;
\For{$i\in\{1...B\}$ (in parallel)}{
\vspace{.5mm}
\texttt{Draw functions:} $t\sim{\cal{T}},t'\sim{\cal{T}}$\;
$\x_i=t({\s}_i)$,~
$\x'_i=t'({\s}_i)$\;
$\z_i=g_{\theta}(f_{\theta}(\x_i));~\z'_i=g_{\xi}(f_{\xi}(\x'_i))$\;
$\uu_i=g_{\xi}(f_{\xi}(\x_i));~\uu'_i=g_{\theta}(f_{\theta}(\x'_i))$\;
}
\Comment{\texttt{Mine views}}
\texttt{Fetch a mini-batch} $\{{\bf c}_j\}_L$ \texttt{from} $\cal{D}$\;
\For{$j\in\{1...L\}$ (in parallel)}{
\vspace{.5mm}
\texttt{Draw function:} $t\sim{\cal{T}}_m$

$\x_{c,j}=t({\bf c}_j);~\y'_{c,j}=f_{\xi}(\x_{c,j})$\;
}
\texttt{Let} $\mathcal{S} = \{\y'_{c,j}\}_{j=1}^L$\;
\For{$i\in\{1...B\}$ (in parallel)}{
\texttt{Draw function:} $t\sim{\cal{T}}_m$\;
\vspace{.5mm}
$\x_{m,i}=t(\s_i);~\y_{m,i}=f_{\theta}(\x_{m,i})$\;
\vspace{.5mm}
\texttt{Find} ${\cal{N}}_k(\y_{m,i})$, \texttt{the} $k$-NNs \texttt{of} $\y_{m,i}$ in $\mathcal{S}$\;
\vspace{.5mm}
\texttt{Randomly select} $\y'_{m,i}$ \texttt{from} ${\cal{N}}_k(\y_{m,i})$\;
$\vv_i=h_{\theta}\left(g_{\theta}\left(\y_{m,i}\right)\right);\vv'_i=h_{\xi}(g_{\xi}(\y'_{m,i}))$\;
}
\Comment{\texttt{Update parameters}}
\vspace{.5mm}
${\cal{L}}=\hspace{-1mm}\sum\limits_i d ( q_{\theta}(\z_i) , \z'_i) + d( q_{\theta}(\uu'_i),\uu_i ) + \lambda d(  r_{\theta}(\vv_i), \vv_i')$\;
\vspace{-.3mm}
$\theta \leftarrow$ \FuncCall{Optimizer}{$\theta, \mathcal{L}/B, \eta$}\;
$\xi \leftarrow \tau \xi + (1-\tau)\theta$\;
}
\vspace{-.2em}
\end{algorithm}

\section{Mining: Implementation details}\label{app:mining-graph}

At a given training iteration, for every sample in the batch, we mine for views in the same pool of candidates of size $L$. Depending on the type of data, the mining for a given sample can be restricted to a subset of that pool of candidates.

{\bf Image datasets.}~When training \alg on images, we use two different dataloaders. The first is the main dataloader that creates batches of size $B$, the second dataloader is independent from the first and is used to sample candidates, and thus has a batch size of $L$. When $L>B$, the second dataloader consumes the dataset before the end of the training epoch, in this case we simply reset the candidate dataloader as many times as needed.

\begin{figure}[t!]
\centering
 \includegraphics[width=0.75\textwidth]{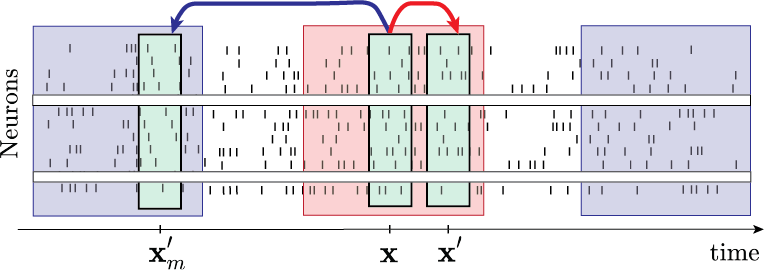}
\caption{ \footnotesize {\em Visualization of augmentations used for neural activity.} Within a small local window around each anchor sample, we consider the nearby samples (red) to be potential positive examples. Outside of a safe zone, we can label more distant samples (blue) as either negative examples (in contrastive learning) or we can also use these points as candidate views to mine from (in \alg). Randomized dropout is illustrated via white bars corresponding to the dropping of the same neurons in all three views.
\vspace{-1mm}
\label{fig:augschematic}}
\end{figure}

{\bf Neural datasets.}~When training \alg on neural datasets, or temporal datasets in general, we restrict mining for a given sample to candidates that are temporally farther in time, as illustrated in Figure~\ref{fig:augschematic}. Implementation-wise, we use a global pool of candidates of size $L$ for simplicity, then when computing the distance matrix used to determine the $k$-nearest neighbors, we mask out the undesired correspondences in the matrix. 

\section{Experimental setup: Image datasets}\label{app:exp-images}

{\em Notation}~~Let MLP$(i, h, o)$ be a linear layer with input size $i$ and output size $h$, followed by batch normalization, rectified linear units (ReLU) and a linear layer of output size $o$. Like in \cite{grill2020bootstrap}, we use these multi-layer perceptrons (MLPs) of depth 2 for projectors and predictors. 


{\bf Architecture.}
We use the CIFAR variant of ResNet-18 as our backbone \cite{chen2020simple}. The representation $\y$ corresponds to the output of the final average pool layer, which has a feature dimension of 512.
We use MLP$(512, 4096, 256)$ for the first projector $g_\theta$ and MLP$(256, 4096, 256)$ for its corresponding predictor $q_\theta$. For the pair of projector/predictor ($h_\theta$/$r_\theta$) dedicated to mined views, we use MLP$(256, 4096, 256)$ and MLP$(256, 4096, 256)$, respectively. 

{\bf Class of transformations.}~
During training, we generate augmented views using the following transformations ($\mathcal{T}$) \cite{chen2020exploring, huang2021selfadaptive}:
\vspace{-0.6em}
\begin{itemize}
    \itemsep-0.2em 
    \item Random cropping: Images are resized to $32$x$32$ using bicubic interpolation, with random area ratio between $0.2$ and $1.0$, and a random aspect ratio between $3/4$ and $4/3$.
    \item Random horizontal flip: the image is fliped left to right with a probability of $0.5$.
    \item Color jittering: the brightness, contrast, saturation, and hue of the image are randomly changed with strengths of $(0.4, 0.4, 0.4, 0.1)$ with a probability of $0.8$. 
    \item Color dropping: the image is converted to gray scale with a probability of $0.1$.
\end{itemize}
\vspace{-0.6em}
When mining, we only use random cropping with a random area ratio between $0.8$ and $1.0$ to augment views ($\mathcal{T}'$).

{\bf Training.}
We use the SGD optimizer with a learning rate of $0.03$, a momentum of $0.9$ and weight decay of $5*10^{-4}$. After a linear warmup period of $10$ epochs, the learning rate is decayed following a cosine decay scheduler.
The exponential moving average parameter $\tau$ is also decayed from $0.996$ to $1.$ following a cosine decay scheduler. We train \alg for $800$ epochs and use a batch size of $B=512$, as well as a pool batch size of $L=512$, and $k=1$. We use a mining weight of $\lambda = 0.1$ linearly ramped-up for $10$ epochs.
\byol is trained using the same relevant hyperparameters. In our experiments, we use the same random seeds for both \alg and \byol.

{\bf Evaluation Protocol:}
Following the evaluation procedures described in \cite{chen2020simple,grill2020bootstrap}, we train a linear classifier on top of the frozen representation of the encoder network and report the accuracy on the test sets (We use the public train/test split for both CIFAR datasets). The linear layer is trained without augmentations for $200$ epochs, with an SGD optimizer with a learning rate of $0.4$ decayed by a factor of $10$ at $140$ and $190$ epochs. 

\section{Experimental details: Neural data} \label{app:exp-neural}

\subsection{Application 1: Decoding movements from motor cortex}

{\bf Details on neural and behavioral datasets in movement decoding task.}
Neural and behavioral data were collected from two rhesus macaque monkeys (Chewie, Mihi). Both individuals performed a standard delayed center-out movement paradigm (reaching experiment). The subjects were seated in a primate chair and
grasped a handle of a custom 2-D planar manipulandum that controlled a computer cursor on a screen. In the first dataset from Chewie, the individual began each trial by moving to a 2 x 2 x 2 cm target in the center of the workspace, and was instructed to hold for 500-1500 ms before another 2 cm target
was randomly displayed in one of eight outer positions regularly spaced at a radial distance of 8
cm. For Mihi, this is followed by another variable delay period of 500 to 1500 ms to plan the movement before an auditory ‘Go’ cue. The sessions with Chewie omitted this instructed delay
period and the ‘Go’ cue was provided when the outer target appeared. Both individuals were required
to reach to the target within 1000-1300 ms and hold within it for 500 ms to receive an auditory
success tone and a liquid reward.

Both individuals were surgically implanted a 100-
electrode array (Blackrock Microsystems, Salt Lake City) in their primary
motor cortex (M1). To record the spiking activity of single neural units, threshold crossings of six times the root-mean square (RMS)
noise on each of the 96 recording channels are initially recorded.  
After each session, the neural waveform data was sorted using Offline Sorter (Plexon, Inc, Dallas, TX) to identify single neurons and discarded all waveforms believed to be multi-unit activity.

Data is only recorded when the primate is performing the reaching task, we note such instance a "trial". We split the trials time-wise, using a $70/10/20$ ratio, to obtain our training, validation and test sets. The temporal splits gives us a better estimate of the prospective prediction compared to a random split \cite{sheridan2013}.
The activity of individual neurons was binned (100 ms intervals) to produce firing rates for roughly 150 neurons across two days.

{\bf Class of transformations.}~
During training, we generate augmented views and mined views using the following transformations ($\mathcal{T} = \mathcal{T'}$):
\vspace{-0.6em}
\begin{itemize}
    \itemsep-0.2em 
    \item Temporal Jitter: a sample within $200$ms is used as a positive example.
    \item Dropout: mask out neurons with a probability uniformly sampled between $0.$ and $0.2$.
    \item Noise: add gaussian noise with standard deviation of 1.5, with a probability of $0.5$.
    \item Pepper or Sparse additive noise: increase the firing rate of a neuron by a $1.5$ constant with a probability of $0.3$. This augmentation is applied on the sample with a probability of $0.5$.
\end{itemize}
\vspace{-0.6em}
Because these datasets correspond to a collection of trials, we restrict mining to candidates that are in different trials from the anchor sample.

{\bf Network Architecture.}~For the encoder, we use an MLP which is $4$ blocks deep. Each block consists of a linear layer with output size $64$ followed by batch normalization (BN) and rectified linear units (ReLU). The final layer has an output size of $32$ and no BN or activation.
We don't use projectors, predictor $q_\theta$ used for augmented views is MLP$(32, 128, 32)$, and predictor $r_\theta$ used for mined views is MLP$(32, 128, 32)$.

{\bf Training.}~We use the AdamW optimizer with a learning rate of $0.02$ and weight decay of $2*10^{-5}$. After a linear warmup period of $100$ epochs, the learning rate is decayed following a cosine decay scheduler.
The exponential moving average parameter $\tau$ is also decayed from $0.98$ to $1.$ following a cosine decay scheduler. We train \alg for $1000$ epochs and use a batch size of $B=512$, as well as a pool batch size of $L=1024$, and $k=5$. We use a mining weight of $\lambda = 1.$ linearly ramped-up for $10$ epochs. \byol is trained using the same relevant hyperparameters.

{\bf Reach direction prediction task.}
The downstream task we use to evaluate the learned representation, is the prediction of the reach direction during movement. There are 8 possible reach direction in total.
Unlike most classification tasks, there is an inherent cyclic ordering between the different classes. Thus, we estimate the angles corresponding to each reach direction, and evaluate their cosine and sine. The linear layer outputs a 2d vector $[x, y]$ that predicts $[\cos{\theta_r}, \sin{\theta_r}]$. We train the network using a mean-squared error loss. Once the network is trained, to readout out the predicted reach direction label, we use the following formula:

\begin{equation}
    l_{\text{predicted}} = \lfloor \frac{4}{\pi} (\text{atan2}(y, x)\ \mathrm{mod}\ 2\pi )\rceil 
\end{equation}

{\bf Evaluation Procedure.}
We train a linear classifier on top of the frozen representation of the encoder network and report the accuracy on the test sets. The linear layer is trained for $100$ epochs using the AdamW optimizer with a learning rate of $0.01$. We sweep over 20 values of the weight decay $\{2^{−10}, 2^{−8}, 2^{−6}, \dots, 2^{6}, 2^{8}, 2^{10}\}$ on the valudation set, and report the accuracies of the best validation hyperparameter on the test set.

\begin{figure}[h] 
\centering
\subfloat[]{
  \includegraphics[width=0.3\textwidth]{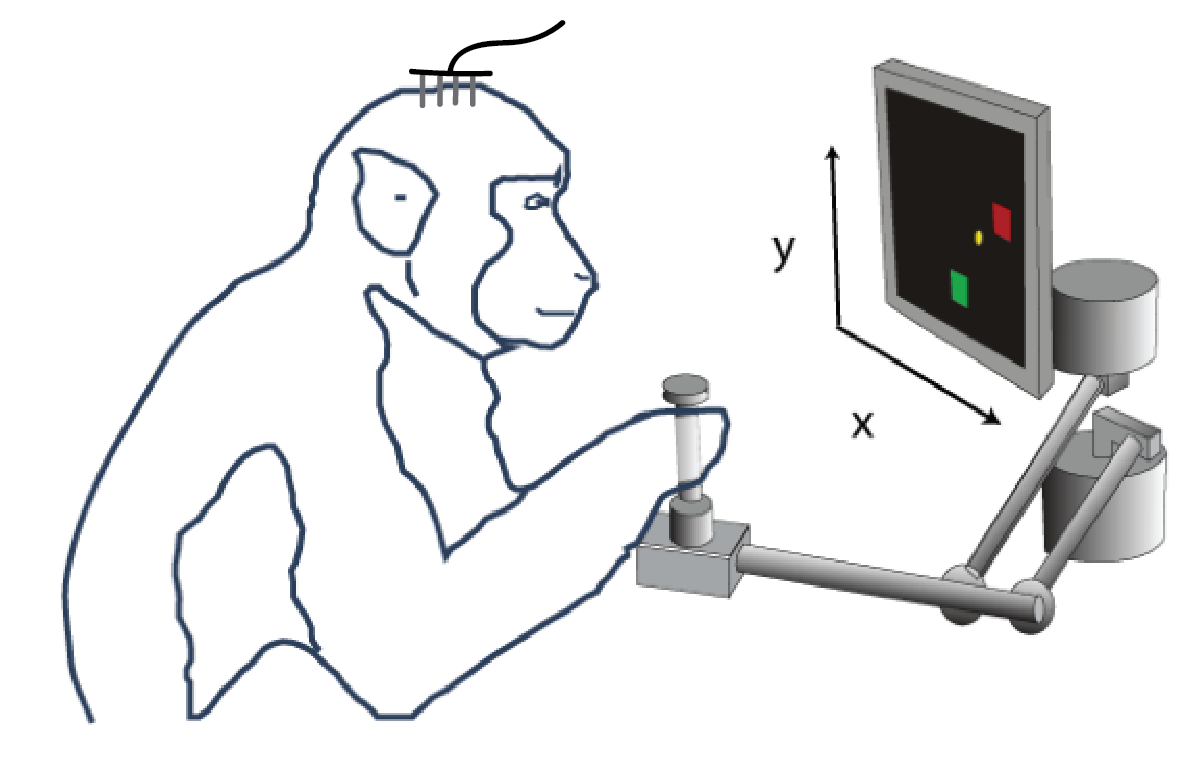}}\ %
  \hspace{5mm}
\subfloat[]{
  \includegraphics[width=0.55\textwidth]{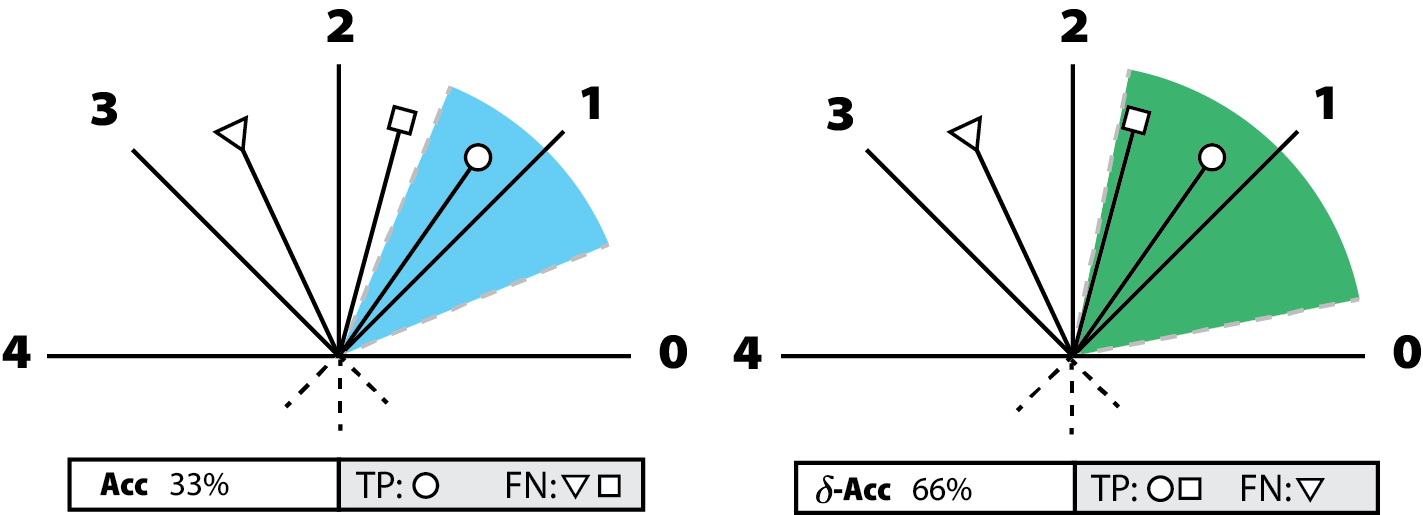}}
\caption{\footnotesize{\em Reach direction prediction task}. (a) Sketch of primate performing reaching task. (b) Illustration depicting how the accuracy and $\delta$-accuracy are computed. The three points have reach direction 1 as their ground truth. TP is true positive and FN is false negative. The highlighted areas correspond to the area a point should fall in to be considered a true positive and be counted towards the corresponding accuracy.}
\label{fig:monkey-task}
\end{figure}

More specifically, we report two different metrics that are computed over the validation set. The Accuracy is the conventional classification accuracy that is obtained when assigning the predicted reach angle to the closest corresponding reach direction. The second metric, \deltacc, is obtained when considering that a prediction is a true positive if it is within a slightly larger window around the true reach direction (an analogy to top-k metrics). (Fig~\ref{fig:monkey-task}-b).

\begin{equation*}
  \begin{split}
    \text{TP}_\text{Acc} = |\frac{4}{\pi} (\text{atan2}(y, x)\ \mathrm{mod}\ 2\pi) - l| < 1
  \end{split}
\quad\quad
  \begin{split}
    \text{TP}_{\delta-\text{Acc}} = |\frac{4}{\pi} (\text{atan2}(y, x)\ \mathrm{mod}\ 2\pi) - l| < 1.5
  \end{split}
\end{equation*}

\subsection{Application 2: Decoding sleep states from rodent cortex}
\label{app:mouse}

{\bf Details on neural and behavioral datasets in arousal state decoding.}
Extracellular single unit spiking was collected from chronically implanted, freely behaving animals. Tetrode arrays were implanted without drives into mouse CA1 (C57BL/6) and rat V1 (Long Evans). Following recovery, neural data were recorded at 25 kHz continuously during free behavior. Raw data were processed and clustered using standard pipelines. Data was bandpassed (500-10,000 Hz) and clustered using MountainSort \cite{mountainsort, spikeinterface}. Single units were identified in the clustering output via XGBoost.

Arousal state was scored using standard polysomnographic methods. Local field potentials (LFP) from 8/64 channels were averaged together, lowpassed (250 Hz), and downsampled. Video (15 fps) was processed using a CNN \cite{deeplabcut} to track animal position and movement. Trained human scorers evaluated the LFP power spectral density and integral of animal movement to evaluate waking, NREM and REM sleep.    

We split the 12 hour block of data temporally using an $70/10/20$ ratio, to obtain our training, validation and test sets. 
The activity of individual neurons was binned ($4$s intervals) to produce firing rates for roughly 40 and 120 neurons from CA1 and V1, respectively.

{\bf Training.}
We use the same hyperparameters as for the monkey datasets, except that the representation size is larger ($64$), and the temporal augmentations are different. With temporal jitter, we consider any two samples that are at most $12$s apart to be positive examples and when mining we restrict the candidates to be at least $30$min before or after the anchor sample.

{\bf Arousal state prediction task.}
We train a linear classifier on top of the frozen representation of the encoder network to predict the arousal state. 
\newpage 
\section{Is \alg worth the extra computational load?}\label{app:compute}

In one iteration, \alg receives $3$ batches worth of views, compared to $2$ for \byol. Thus, there is a possibility that \alg performs better than \byol simply because of the higher effective batch size used during training. To rule this possibility out, we try both training \byol for $50\%$ more epochs and training \byol using a $50\%$ bigger batch size, and report the results in Table~\ref{tab:adjusted-byol}. We show that the improvements we find through with \alg go beyond extra training time.

\begin{table}[h]
\centering
\caption{\footnotesize{\em Training \byol with adjusted batch size and number of epochs.} We report the linear evaluation accuracies on CIFAR-10 using ResNet-18. \label{tab:adjusted-byol}}
 \vspace{0.1in}
\begin{tabular}{lccc}
\hline
     & Batch size & Number of epochs & Accuracy \\ \hline
\byol & 512        & 800              & 91.71    \rule{0pt}{2ex}\\
\byol & 512        & 1200             & 91.75    \\
\byol & 768        & 800              & 91.65    \\ \hline
\alg & 512        & 800              & 92.10    \rule{0pt}{2ex}\\ \hline
\end{tabular}
\end{table}

\begin{figure*}[h]
\centering
   \includegraphics[width=0.6\textwidth]{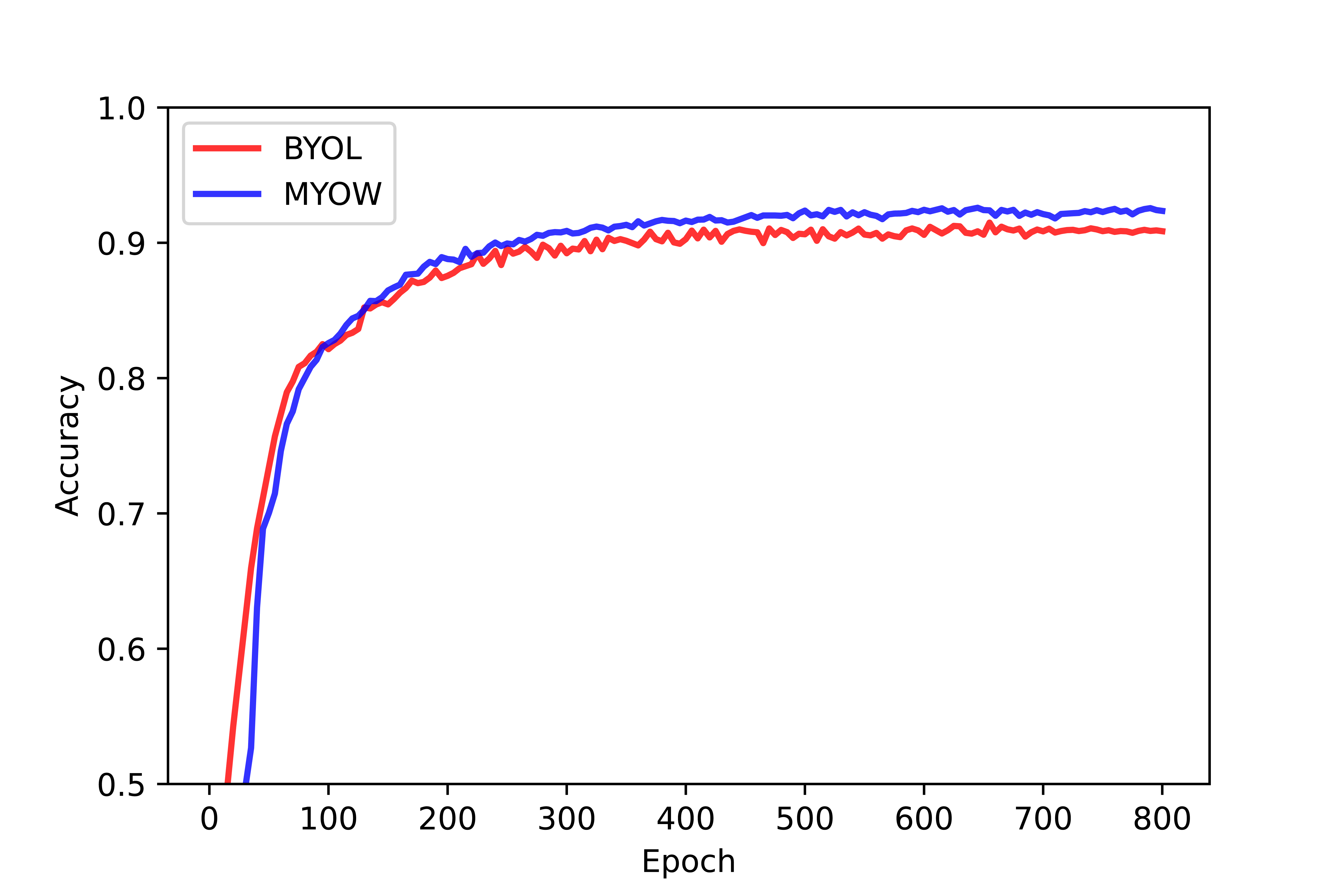}
   \caption{\footnotesize{\em Accuracy under linear evaluation, CIFAR10, ResNet18}. \byol (bottom), \alg (top).}
   \label{fig:acc_curve}
\end{figure*}

When we examine the accuracy curves during training (Figure~\ref{fig:acc_curve}), we find that \alg surpasses the final accuracy of \byol after only $300$ epochs of training. Thus, in the case of this dataset, we can justify the extra computational load that comes with using \alg, as it yields better results early on in training.

\section{What makes for good mined views?}\label{app:minequality}

In Table~\ref{tab:minespace}, we compare the outcomes of using the online representations of the candidates compared to their target representations when looking for the $k$-nn of the online representation of the anchor sample. We find that both strategies yield similar results while mining in the target is less computationally expensive. 

\begin{table}[h]
\centering
\caption{\footnotesize{\em Mining in online versus. target space.} We report the linear evaluation accuracies on CIFAR-10 using ResNet-18, as well as an approximation of the computational load factor with \byol as the baseline.  \label{tab:minespace}}
 \vspace{0.1in}
\begin{tabular}{lccc}
\hline
     & Mining in & Computational factor & Accuracy \rule{0pt}{2ex} \\ \hline
\byol &   -      & 1.00   & 91.71  \rule{0pt}{2ex} \\
\alg & online        & 1.75  & 92.13 \\
\alg & target        & 1.50   & 92.10   \\ \hline
\end{tabular}
\end{table}

\begin{figure*}
\centering
   \includegraphics[width=1.0\textwidth]{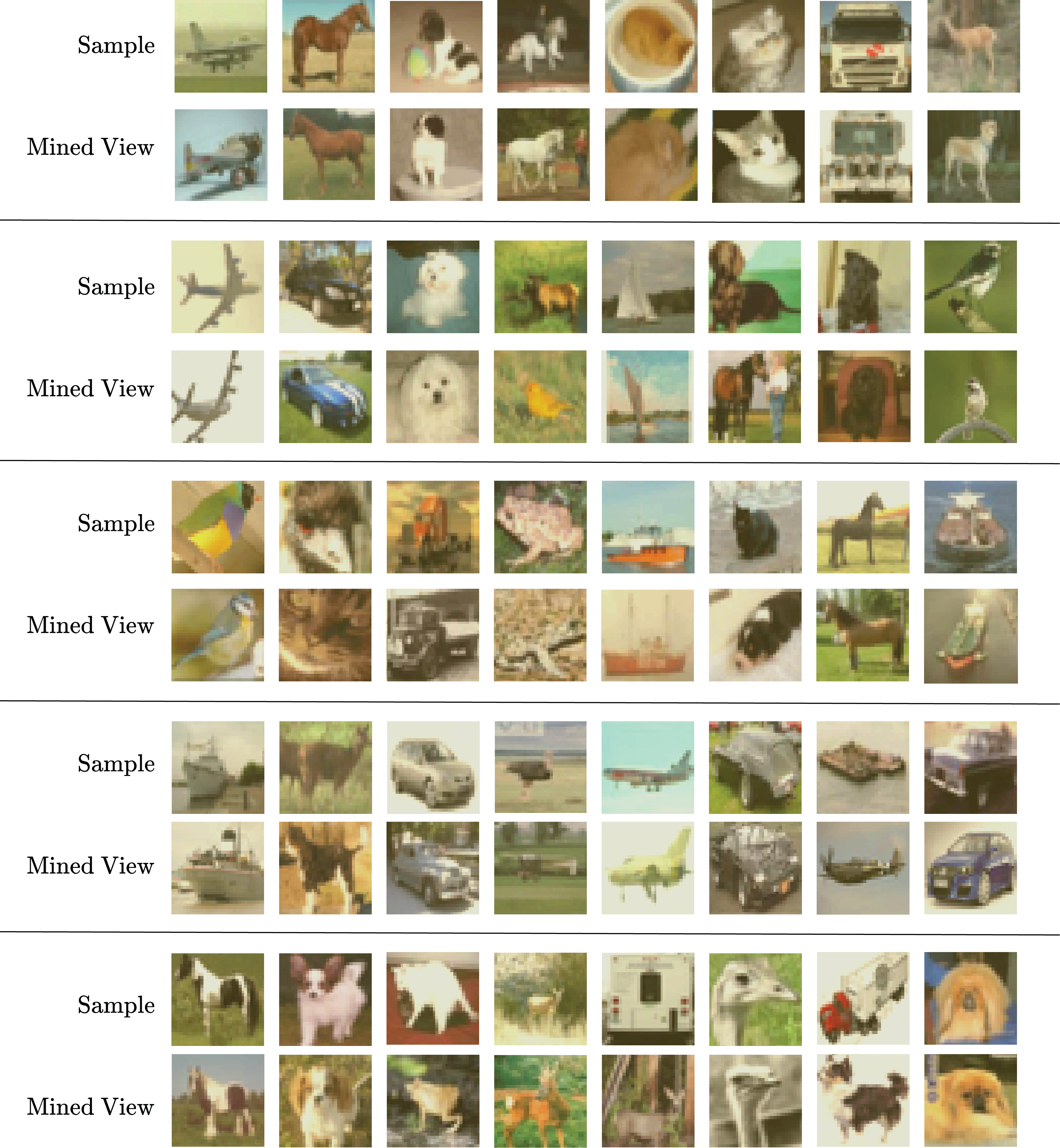}
   \vspace{2mm}
   \caption{\footnotesize{\em Examples of views mined by \alg}. We visualize the views mined by \alg during training on the CIFAR-10 dataset at epoch $400$.}
   \label{fig:moreimages}
   \vspace{-3mm}
\end{figure*}

We analyse the views that are being mined when training \alg on CIFAR-10. In Figure~\ref{fig:moreimages}, we show a random collection of views paired during mining. 

\begin{figure*}[h]
\centering
   \includegraphics[width=0.6\textwidth]{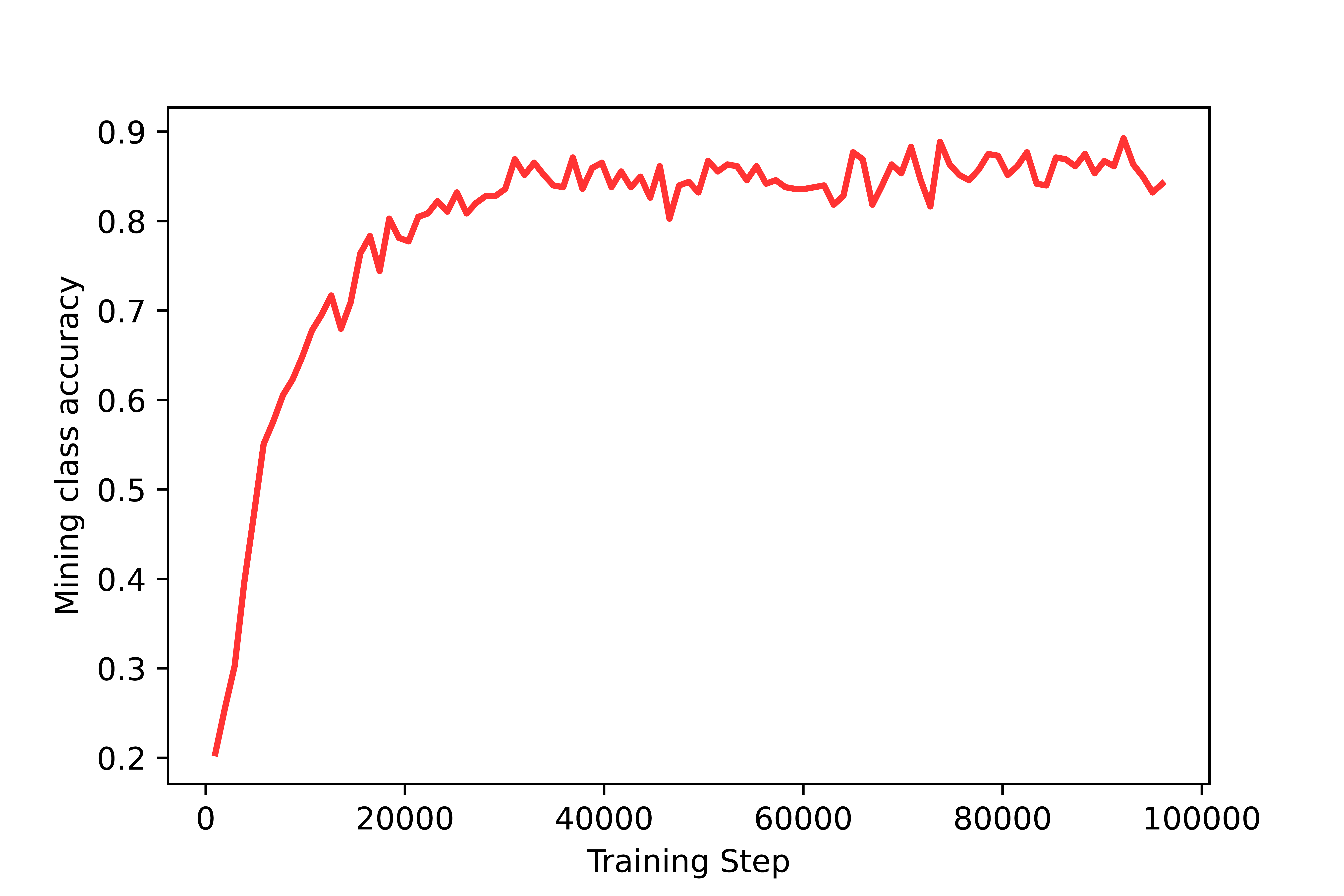}
   \caption{\footnotesize{\em Mining class accuracy during training.} This metric is reported on CIFAR-10 using ResNet-18.}
   \label{fig:mine_acc_curve}
\end{figure*}

\begin{figure*}[h]
\centering
   \includegraphics[width=0.45\textwidth]{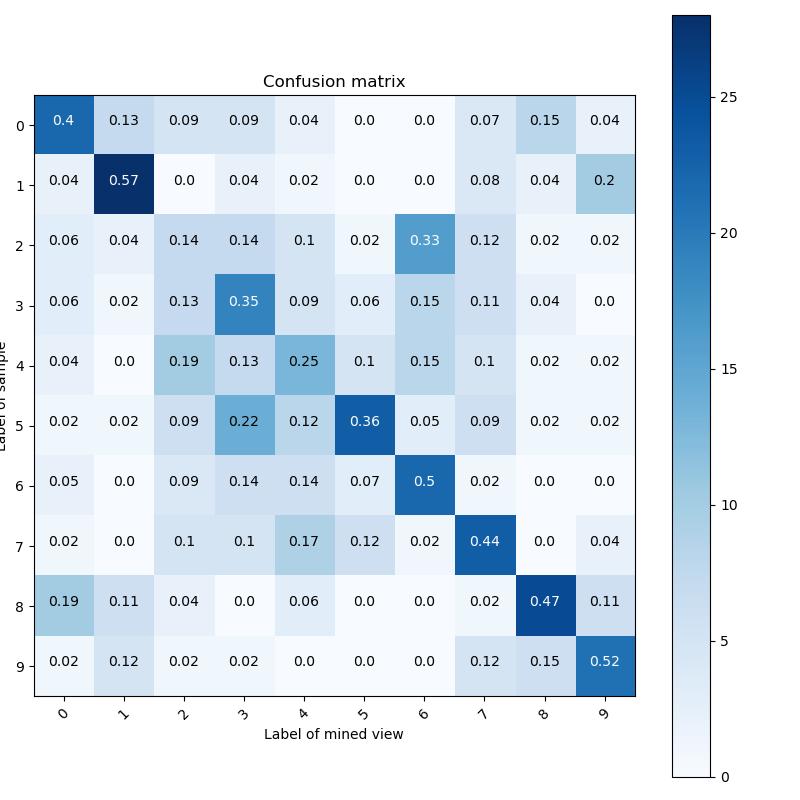}
   \includegraphics[width=0.45\textwidth]{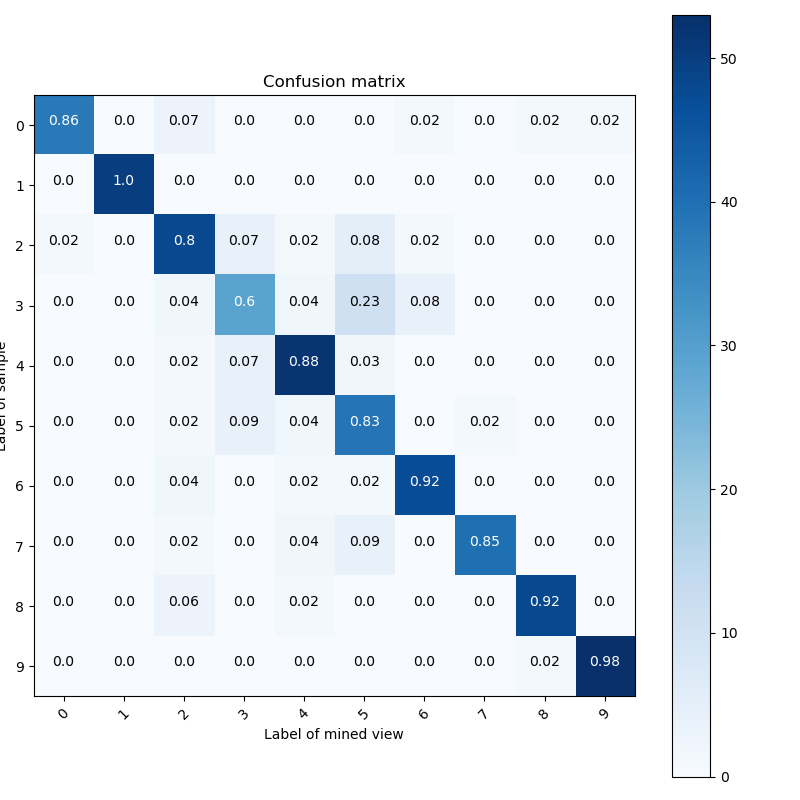}
   \caption{\footnotesize{\em Mining class confusion matrices at different stages of learning}. We compute the confusion matrix at epochs $100$ (right) and ($700$) when training on CIFAR-10.}
   \label{fig:cm_cifar10}
\end{figure*}

\alg relies on mining views that are semantically similar, but it is not clear how robust \alg is to ``bad'' mined views. While we are not able to give a definitive answer to this question, we find that even when certain mined views have a different class from the anchor samples, \alg still yields competitive results. In Figure~\ref{fig:mine_acc_curve}, we look at the mining class accuracy, defined as the percentage of mined views that share the same class as their anchor samples, and find that the accuracy steadily increases during training and that the relatively low accuracy at the beginning of training does not hinder the performance of \alg. The mining class accuracy gives us a better understanding of the mining, but it is not a reflection of the goodness of the mining, as we do not know what makes for a good mined view and whether a inter-class mined views could be ``good''.
We also visualize, in Figure~\ref{fig:cm_cifar10}, the mining class confusion matrices at epochs $100$ and $700$ of training.

\section{Ablation on the projector}\label{app:projector}

In Table~\ref{tab:projectors-full} and Table~\ref{tab:projectors-cifar}, we report the results of \alg on the MNIST and CIFAR-10 datasets for different architectures used for incorporating mined views into our objective: cascaded projectors (used in \alg), parallel projectors and single projector For MNIST, we show the results for two different settings, weak augmentation (Crop only) and strong augmentation (All). Overall, we find that separating the projection spaces for augmented and mined views is better, with the cascading yielding the best results.

\begin{figure}
\begin{minipage}{.45\textwidth}
  \centering
  \captionof{table}{\footnotesize{\em Comparing different projector architectures for incorporating mined views.} MNIST classification accuracy (in \%) with \alg for different architectures. \vspace{0.1em} \label{tab:projectors-full}}
  \begin{tabular}{llcc}
\hline
Arch     & Dimension & \multicolumn{2}{c}{MNIST} \rule{0pt}{2ex}\\ 
         &           & Crop only   & All \\ \hline
Cascaded & 16        & {\bf 99.20} & {\bf 99.33}\rule{0pt}{2ex} \\
Cascaded & 128       &  98.09 & 98.80 \\
Parallel & 16        &  96.33 & 98.71 \\
Parallel & 128       &  97.75 & 98.12 \\
Single & 16          &  97.13 & 97.48 \\
Single & 128         &  98.75 & 98.31 \\
\hline
\end{tabular}
\end{minipage}%
\hspace{4em}
\begin{minipage}{.4\textwidth}
  \centering
    \captionof{table}{\footnotesize{\em Comparing different projector architectures for incorporating mined views.} CIFAR-10 classification accuracy (in \%) with \alg for different architectures. \vspace{0.1em} \label{tab:projectors-cifar}}
  \begin{tabular}{llcc}
\hline
Arch      & CIFAR-10 \rule{0pt}{2ex} \\ \hline
Cascaded projectors &{\bf 92.10}\rule{0pt}{2ex} \\
Parallel projectors &  92.01 \\
Single projector &  91.84 \\
\hline
\end{tabular}
\end{minipage}
\end{figure}


\section{Ablation on the class of transformations}\label{app:trans}

We study how the choice of the set of transformations used in the mining process, impacts the quality of the representation. In Table~\ref{tab:class_t_mine}, we report the accuracies under linear evaluation when we use different classes of transformation $\mathcal{T}'$.

\begin{table}[!h]
\caption{{\em Class of transformation for mined views.} We report the accuracies under linear evaluation of \alg trained on CIFAR-10 using ResNet-18, for different classes of transformation $\mathcal{T}'$ \label{tab:class_t_mine} \vspace{0.5em}}

\centering
\begin{tabular}{ccccc}
\hline
Crop & Flip & Color jitter & CIFAR-10 \rule{0pt}{2ex}\\ \hline
\checkmark $(0.2-1.0)$ & \checkmark & \checkmark & 91.63 \rule{0pt}{2ex} \\
\checkmark $(0.8-1.0)$  &  & & 92.10  \\
\xmark &  & & 92.08  \\\hline
\end{tabular}
\end{table}

\section{Gaining insights into across-sample prediction}
\label{sec:dsprites}
Based upon our experiments on neural data, we conjectured that the diversity introduced by \alg makes it possible to learn effectively, even when the augmentations provided to the network are too local to drive learning in \byol. We thus designed an experiment using the dSprites dataset \cite{dsprites17}, as it allows control over the generation of data over multiple latent positions. 

The dSprites dataset is comprised of a total of 737,280 images. Each image has an associated shape, orientation, scale and 2D position. Each one of these latent variables has a finite number of possible values because of the procedural nature of the dataset.
To generate the downsampled training sets used in our experiment, we uniformly sample $50\%$ of the orientation latent values as well as $50\%$ of the scale latent values, and only consider the corresponding images, thus effectively creating holes in the latent manifold. 
The dataset is further downsampled at a given rate $r$ to generate the train set, the remaining images form the test set. The size of the train set is effectively $0.25 * r$ that of the entire dataset. In our experiment, we generate training sets that are $30\%$, $15\%$ and $7.5\%$ the size of the dataset.

\begin{figure}[!h]
\centering
\includegraphics[width=\textwidth]{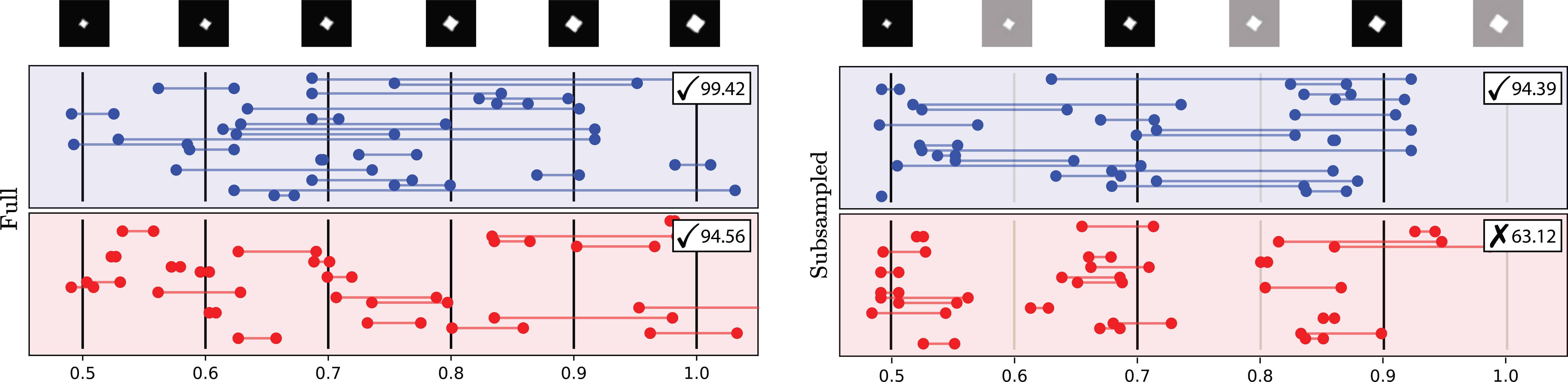}
\caption{\footnotesize {\em Understanding predictive learning when augmentations are too local.} Each segment represents a pair of views (red for augmented, blue for mined) of the corresponding latent scale (x-axis). The vertical lines represent the original scales of samples pre-augmentation. We examine the case where we have access to the full dataset (left) and when we have only half of the latent positions (3/6) and 7.5\% of the remaining samples (right).}
\label{fig:dsprites}
\end{figure}

When we train \byol and \alg on a sufficiently dense sampling of the latent positions (30\%), we observe that both models can classify on unseen latent positions with nearly 100\% accuracy (Figure~\ref{fig:dsprites}). However, when we consider the undersampled condition (7.5\%), \byol fails to generalize to the unseen positions, resulting in a low accuracy of around 60\%. In contrast, \alg maintains a high accuracy of 94\% despite the limited training data. These findings suggest that in settings where the data manifold is sparsely sampled, \alg provides a way to build predictions across different but similar data samples.

\section{Augmentations for spiking neural data}\label{app:augmentations}

{\bf Temporal jitter.}
As in previous work in temporal contrastive learning \cite{oord2018representation,sermanet2018timecontrastive,dwibedi2019learning,Le_Khac_2020,banville2020uncovering}, we can use nearby samples as positive examples for one another.

{\bf Randomized dropout.}
When working with neural data, we consider randomized dropout \cite{bouthillier2016dropout} as an augmentation. The dropout rate is uniformly sampled between $p_{\text{min}}$ and $p_{\text{max}}$.

{\bf Gaussian noise.}
Random Gaussian noise with mean $0$ and standard deviation $1.5$ is applied before normalizing the firing rates.

{\bf Random pepper.}
In contrast to dropout, applying random pepper consists of randomly activating neurons. Similar to the dropout probability, a pepper probability is used to specify the probability of activating a neuron. The activation consists in adding a constant to the firing rate.

In Table~\ref{appendix:monkeyaug}, we show how different augmentations impact neural datasets not detailed in the main text. The findings are echoed through all monkey datasets.

\begin{table}[h]
\caption{{\em How  augmentations impact our ability to decode movements accurately.} To understand how different augmentations impact the representations obtained with \byol and \alg for all four datasets, we computed the Accuracy in our reach direction prediction task when we apply a given set of transformations.\label{appendix:monkeyaug}}
\vspace{0.4em} 
\centering
\resizebox{0.8\textwidth}{!}{
\begin{tabular}{lcccccccc}
\hline
                     & TJ & Drop & Noise & Pepper & \multicolumn{4}{c}{Accuracy}                                  \\
                     &    &       &       &        & Chewie-1 & Chewie-2 & Mihi-1 & Mihi-2 \\ \hline
\byol& \checkmark &            &            &            & 41.75 & 40.83 & 43.98 & 44.10 \\ 
     &            & \checkmark & \checkmark & \checkmark & 55.70 & 49.37 & 47.61 & 43.12 \\
     & \checkmark & \checkmark &            &            & 61.39 & 56.48 & 59.53 & 58.37 \\
     & \checkmark & \checkmark & \checkmark & \checkmark & 63.80 & 57.17 & 59.50 & 60.82 \\ \hline
\alg & \checkmark &            &            &            & 46.61 & 42.91 & 42.08 & 44.13 \\ 
     &            & \checkmark & \checkmark & \checkmark & 53.15 & 46.17 & 51.44 & 48.72 \\
     & \checkmark & \checkmark &            &            & 67.97 & 58.21 & 68.93 & 63.90 \\
     & \checkmark & \checkmark & \checkmark & \checkmark & 70.41 & 60.95 & 70.48 & 64.35 \\\hline
\end{tabular}}
   \end{table}

In Table~\ref{appendix:mouseaug}, we show the impact of both temporal shift and dropout on the performance on rodent datasets. Here, we also find that both components are important to achieving good performance.

\begin{figure}[t]
\centering
\captionsetup{type=table} 

\begin{minipage}[b]{0.8\textwidth}
\caption{{\em How  augmentations impact our ability to decode sleep and wake states accurately.} To understand how different augmentations impact the representations obtained with \byol and \alg for the two datasets labeled {Sleep}, we computed the F1-score for different classes of augmentations in two brain areas. \label{appendix:mouseaug} \vspace{0.5em}}

\centering
\begin{tabular}{lcc|cc}
\hline
                     & TS & RDrop  & \multicolumn{2}{c}{F1-score}\\
                     &    &          & Rat-V1 & \hspace{3mm}Mouse-CA1  \\ \hline
\byol &  \checkmark &       & 68.66 & 87.73 \\
      &   & \checkmark     & 79.31 & 88.84 \\
& \checkmark & \checkmark  & 85.42 & 93.24 \\ \hline
      
\alg  & \checkmark &     & 72.13 & 90.01 \\
      &  & \checkmark     & 85.60 & 83.33 \\
      & \checkmark & \checkmark  & 88.01 & 93.70 \\\hline      
\end{tabular}
\end{minipage}
\end{figure}

\newpage
\section{Visualization of the latent neural space}\label{app:latent-neural}

\begin{figure*}[!h]
\centering
   \includegraphics[width=0.7\textwidth]{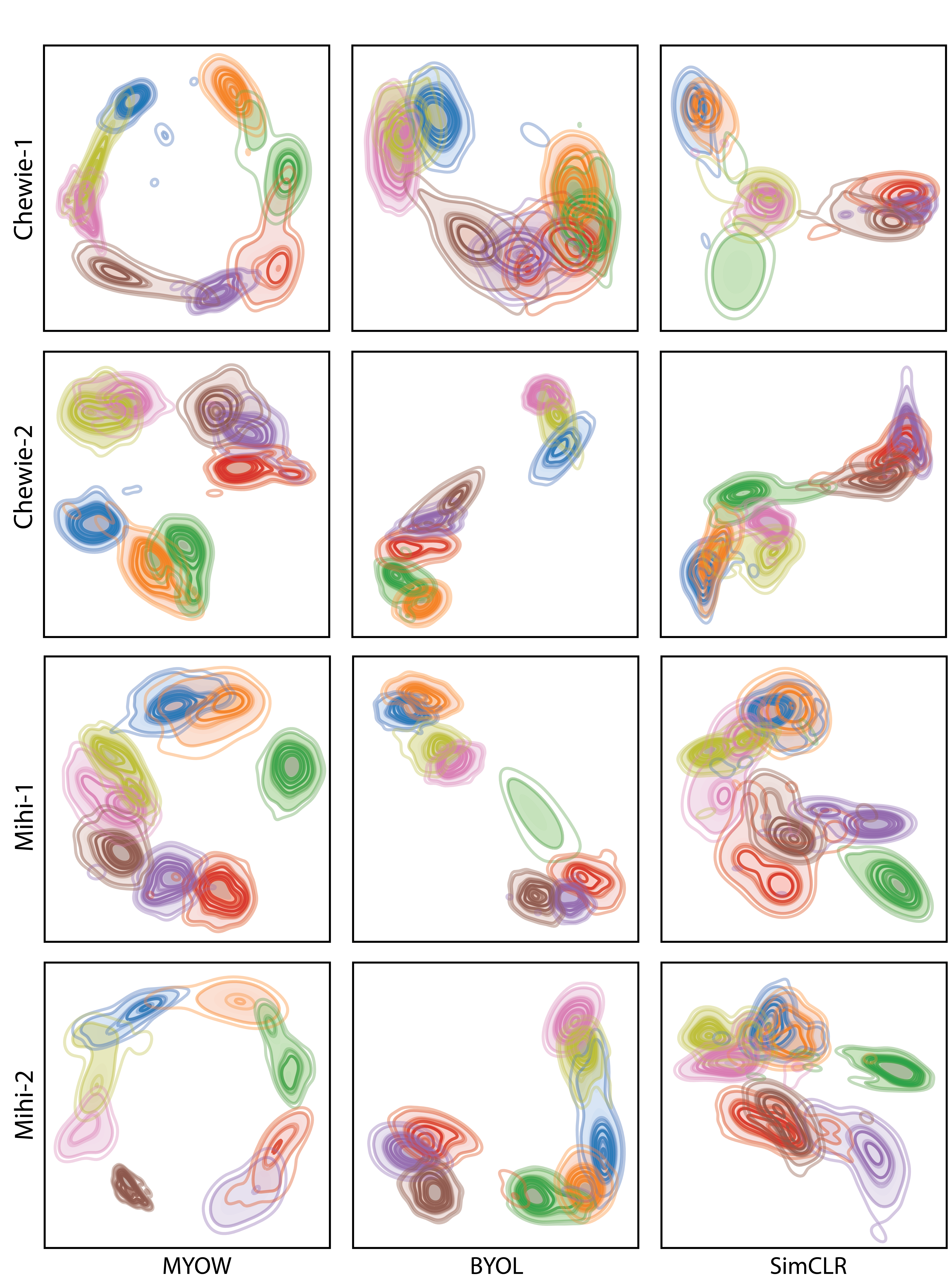}
   \vspace{2mm}
   \caption{\footnotesize{\em Visualization of the learned representation}. Using t-SNE, we visualize the representation spaces when training \alg, \byol and \simclr on all four monkey reach datasets.}
   \label{fig:moreneuro}
   \vspace{-3mm}
\end{figure*}

In Figure~\ref{fig:moreneuro}, we provide the visualizations of the latent spaces for all four monkey reach dataset and can identify a common pattern in the structure uncovered by the different methods.

\end{document}